\documentclass{article}

\usepackage[preprint]{jmlr2e}

\usepackage[numbers]{natbib}
    \renewcommand{\cite}[1]{\citep{#1}}

\usepackage[utf8]{inputenc} 
\usepackage[T1]{fontenc}    
\usepackage{hyperref}       
\usepackage{url}            
\usepackage{booktabs}       
\usepackage{amsfonts}       
\usepackage{nicefrac}       
\usepackage{microtype}      
\usepackage{xcolor}         

\title{Improved Regret of Linear Ensemble Sampling}

\author{\name Harin Lee \email harinboy@snu.ac.kr \\
    \addr Seoul National University
    \AND
    \name Min-hwan Oh \email minoh@snu.ac.kr \\
    \addr Seoul National University
}

\usepackage{enumitem}
\usepackage{amsmath}
\usepackage{amssymb}
\usepackage{amsthm}
\theoremstyle{definition}
\newtheorem{remark}{Remark}
\theoremstyle{plain}
\usepackage{dsfont}
\usepackage{algorithm}
\usepackage{algpseudocode}
\usepackage{paracol}
\usepackage{multicol}
\usepackage{cleveref}
\usepackage{booktabs}
\usepackage[flushleft]{threeparttable} 
\crefname{theorem}{Theorem}{Theorems}
\crefname{corollary}{Corollary}{Corollaries}
\crefname{lemma}{Lemma}{Lemmas}
\crefname{proposition}{Proposition}{Propositions}
\crefname{claim}{Claim}{Claims}
\crefname{remark}{Remark}{Remarks}
\crefname{section}{Section}{Sections}
\crefname{appendix}{Appendix}{Appendices}
\crefname{fact}{Fact}{Facts}
\crefname{algorithm}{Algorithm}{Algorithms}
\crefname{table}{Table}{Tables}

\usepackage{abbreviations}

\begin{document}

\maketitle

\begin{abstract}
In this work, we close the fundamental gap of theory and practice by providing an improved regret bound for linear ensemble sampling. We prove that with an ensemble size logarithmic in \(T\), linear ensemble sampling can achieve a frequentist regret bound of \(\tilde{\Ocal}(d^{3/2}\sqrt{T})\), matching state-of-the-art results for randomized linear bandit algorithms, where $d$ and $T$ are the dimension of the parameter and the time horizon respectively. Our approach introduces a general regret analysis framework for linear bandit algorithms. Additionally, we reveal a significant relationship between linear ensemble sampling and Linear Perturbed-History Exploration (LinPHE), showing that LinPHE is a special case of linear ensemble sampling when the ensemble size equals \(T\). This insight allows our analysis framework to derive a regret bound of \(\tilde{\Ocal}(d^{3/2}\sqrt{T})\) for LinPHE, independent of the number of arms. Our techniques advance the theoretical foundation of ensemble sampling, bringing its regret bounds in line with the best known bounds for other randomized exploration algorithms.
\end{abstract}

\section{Introduction}
\label{sec:introduction}

Ensemble sampling~\citep{lu2017ensemble} has emerged as an empirically effective randomized exploration technique in various online decision-making problems, such as online recommendation~\citep{lu2018efficient,zhu2023deep,zhou2024stochastic} and deep reinforcement learning~\citep{osband2016deep,osband2018randomized,osband2019deep}. 
Despite its popularity, the theoretical understanding of ensemble sampling has lagged behind, even for the linear bandit problem, with previous results revealing sub-optimal outcomes.
For instance, a prior work~\citep{qin2022analysis} demonstrated that linear ensemble sampling could achieve \(\Ocal(\sqrt{T})\) \textit{Bayesian} regret with an ensemble size growing at least linearly with \(T\). 
However, the requirement for the ensemble size to be linear in \(T\) is highly unfavorable and prohibitive in many practical settings. 
A recent work~\citep{janz2024ensemble} showed that a symmetrized version of linear ensemble sampling could provide an improvement in dependence on ensemble size of $\Theta(d \log T)$ and show a frequentist regret bound of \(\tilde{\Ocal}(d^{5/2}\sqrt{T})\). 
However, this regret bound clearly falls short of the existing frequentist regret achieved by standard randomized algorithms such as Thompson Sampling~(TS)~\citep{agrawal2013thompson, abeille2017Linear}
and Perturbed-History Exploration (PHE)~\citep{kveton2019perturbed,kveton2020perturbed, kveton2020randomized}.\footnote{LinTS~\citep{agrawal2013thompson, abeille2017Linear} has regret of $\tilde{\Ocal} \big( \min(d^{3/2} \sqrt{T}, d \sqrt{ T \log K }) \big)$, and LinPHE~\citep{kveton2020perturbed} has regret of  $\tilde{\Ocal} ( d \sqrt{ T \log K } )$. For exchanges between factors of $\Ocal(\sqrt{d})$ and $\Ocal(\sqrt{\log K})$, we refer to the analysis of \citet{agrawal2013thompson}.
Therefore, the existing result for linear ensemble sampling~\citep{janz2024ensemble} has a gap of at least $\Ocal(d)$ compared to LinTS and LinPHE. Besides this gap in regret bounds, there appears to be rather counter-intuitive dependence on ensemble size in  \citet{janz2024ensemble} 
(see \cref{remark:janz_ensemble_size_increasing} in Section \ref{sec:regret bound of linear ensemble sampling}).
}

In this work, we close this fundamental gap by providing an improved regret bound for linear ensemble sampling. 
We prove that linear ensemble sampling with an ensemble size logarithmic in~\(T\) can still attain a frequentist regret bound of \(\tilde{\Ocal}(d^{3/2}\sqrt{T})\), marking the first time that linear ensemble sampling achieves a state-of-the-art result for randomized linear bandit algorithms. 
Our approach not only improves upon the regret bound but also simplifies the algorithm by avoiding the use of symmetrized perturbations, making it more practical for implementation. 
For regret analysis, we present a general, concise framework for analyzing linear bandit algorithms, which may be of independent interest. 
Furthermore, we rigorously reveal the significant relationship between ensemble sampling and PHE for the first time, 
showing that in the regime where the ensemble size equals \(T\), linear PHE (LinPHE) is a special case of linear ensemble sampling. 
With this new insight, we can use the regret analysis for ensemble sampling to derive a regret bound of \(\tilde{\Ocal}(d^{3/2}\sqrt{T})\) for LinPHE, which is independent of the number of arms $K$.

Our main contributions are summarized as follows:

\begin{itemize}[leftmargin=*]    
    \item We prove a $\tilde{\Ocal}(d^{3/2}\sqrt{T})$ regret bound (\cref{thm:enssamp}) for linear ensemble sampling with an ensemble size of $m = \Omega(K \log T)$, where $K$ denotes the number of arms.
    Importantly, our regret bound does not depend on $K$ or $m$ even logarithmically. 
    Our result is the first to establish
    $\tilde{\Ocal}(d^{3/2}\sqrt{T})$ regret for linear ensemble sampling with an ensemble size sublinear in $T$, improving the previous bound by the factor $d$ while maintaining the ensemble size to be logarithmic in $T$.
    
    \item As part of the regret analysis, we present a general regret analysis framework (\cref{thm:general}) for linear bandit algorithms. This framework not only generalizes the regret analysis of randomized algorithms such as ensemble sampling and PHE but also applies to other optimism-based deterministic algorithms. This result can be of independent interest beyond ensemble sampling.
    
    \item 
    We rigorously investigate the relationship between linear ensemble sampling and LinPHE. 
    We show that in the regime of ensemble size $m = T$, LinPHE is a special case of the linear ensemble sampling algorithm. 
    To our best knowledge, this is the first result to show the equivalence between linear ensemble sampling and LinPHE.

    \item As a byproduct, with this new insight into the relationship between linear ensemble sampling and LinPHE, we provide an alternate analysis for LinPHE as an extension of the analysis for linear ensemble sampling, achieving a $\tilde{\Ocal}(d^{3/2}\sqrt{T})$ regret bound with no dependence on $K$.
\end{itemize}

\subsection{Related Work}

The stochastic linear bandit problem~\citep{auer2002using,abbasi2011improved,lattimore2020bandit} is a foundational sequential decision-making problem and a core model for multi-armed bandits with features. Numerous algorithms have been developed for this problem, including deterministic approaches such as UCB-based methods~\citep{auer2002using,chu2011contextual,abbasi2011improved} and randomized algorithms such as Thompson sampling~\citep{thompson1933likelihood,chapelle2011empirical,agrawal2013thompson,abeille2017Linear} and PHE~\citep{kveton2019perturbed,kveton2020perturbed,kveton2020randomized}.

Thompson sampling~\citep{thompson1933likelihood}, a classical randomized method, utilizes the posterior distribution of hidden parameters based on observed data. Initially proposed for Bayesian settings~\citep{russo2014learning,russo2018tutorial}, it has also demonstrated strong performance in frequentist settings~\citep{agrawal2012analysis,agrawal2013thompson,abeille2017Linear}. For the stochastic linear bandit, \citet{agrawal2013thompson} showed that Thompson sampling with a Gaussian prior achieves a regret bound of $\tilde{\Ocal}(d^{3/2}\sqrt{T})$, which can be reduced to $\tilde{\Ocal}(d\sqrt{T \log K})$ for small $K$. However, applying Thompson sampling to more complex problems remains challenging, especially when posterior computation becomes intractable, though approximate methods have been proposed~\citep{lu2017ensemble,wan2023multiplier}.

PHE~\citep{kveton2019perturbed,kveton2020perturbed,kveton2020randomized} is another class of randomized algorithms that does not rely on posterior distributions, making it potentially applicable to more complex settings. In the finite-armed linear bandit model, PHE achieves a $\tilde{\Ocal}(d \sqrt{T \log K})$ regret bound, matching the performance of Thompson sampling for finite arms~\citep{agrawal2013thompson}. However, the relationship between PHE and ensemble sampling remains unexplored in previous studies.

Ensemble sampling~\citep{lu2017ensemble} has gained popularity as a randomized exploration method across various decision-making tasks~\citep{lu2018efficient,zhu2023deep,zhou2024stochastic,osband2016deep,osband2018randomized,osband2019deep}. Despite its empirical success, its theoretical foundation, particularly for linear bandits, is still relatively underdeveloped. \citet{qin2022analysis} showed that linear ensemble sampling achieves \(\Ocal(\sqrt{T})\) Bayesian regret but requires an impractically large ensemble size that scales linearly with \(T\). More recently, \citet{janz2024ensemble} reduced the dependence on ensemble size to $\Theta(d \log T)$ and achieved a frequentist regret bound of \(\tilde{\Ocal}(d^{5/2}\sqrt{T})\). However, the frequentist regret bound of \(\tilde{\Ocal}(d^{3/2}\sqrt{T})\) for ensemble sampling has yet to be achieved.

\section{Preliminaries}

\subsection{Notations}
$\NN$ denotes the set of natural numbers starting from $1$. 
For a positive integer $M$, $[M]$ denotes the set $\{1, 2, \ldots, M\}$.
$\boldsymbol{0}_d$ denotes the zero vector in $\RR^d$ and $I_d$ denotes the identity matrix in $\RR^{d \times d}$. 
We define and work within a probability space $\left( \Omega, \Fcal, \PP \right)$, where $\Omega$ is the sample space, $\Fcal$ is the event set, and $\PP$ is the probability measure.
$\Ncal(\mu, \Sigma)$ denotes the uni- or multi-variate Gaussian distribution with mean $\mu$ and covariance $\Sigma$.
$\land$ denotes logical conjunction (``and'') and $\lor$ denotes logical disjunction (``or'').
With slight abuse of notation, we write $\left\{ \omega \in \Omega : A \right\}$ and $A$ interchangeably when $A$ is some condition, for simplicity.
$\Ocal( \cdot)$ denotes the asymptotic growth rate with respect to problem parameters $d$, $T$, and $K$. 
$\tilde{\Ocal}(\cdot)$ further hides logarithmic factors of $T$ and $d$.

\subsection{Problem Setting}
\label{sec:problem setting}
We consider the stochastic linear bandit problem.
The learning agent is presented with a non-empty arm set $\Xcal \subset \RR^d$.
For $T$ time steps, where $T \in \NN$ is the time horizon, the agent selects an arm $X_t \in \Xcal$ and receives a real-valued reward $Y_t$, where the reward is generated based on a hidden true parameter vector, $\theta^* \in \RR^d$.
Specifically, $Y_t$ is defined as follows:
\begin{equation*}
    Y_t = X_t^\top \theta^* + \eta_t
    \, ,
\end{equation*}
where $\eta_t$ is a zero-mean random noise.
The objective of the agent is to maximize the cumulative reward, or equivalently, to minimize the cumulative regret $R(T)$ defined as
\begin{equation*}
    R(T) := \sum_{t=1}^T \left( \sup_{x \in \Xcal} x^\top \theta^* - X_t^\top \theta^* \right)
    \, .
\end{equation*}

\section{Ensemble Sampling for Linear Bandits}

\begin{algorithm}[t!]
\caption{Linear Ensemble Sampling}
\label{alg:enssamp}
\begin{algorithmic}
    \State \textbf{Input} : regularization parameter $\lambda > 0$, ensemble size $m \in \NN$, 
    \State \qquad initial perturbation distribution $\Pcal_I$ on $\RR^d$, reward-perturbation distribution $\Pcal_R$ on $\RR$,
    \State \qquad ensemble sampling distribution $\{\Jcal_t\}_{t=1}^T$ on $[m]$
    \State Sample $W^j \sim \Pcal_I$ for each $j \in [m]$.
    \State Initialize $V_0 = \lambda I_d$, $S_0^j = W^j$, $\theta_0^j = V_0^{-1} S_0^j$ for each $j \in [m]$
    \For{$t  = 1, 2, \ldots T$}
        \State Sample $j_t \sim \Jcal_t$
        \State Pull arm $X_t = \argmax_{x \in \Xcal} x^\top \theta_{t-1}^{j_t}$ and observe $Y_t$
        \State Update $V_t = V_{t-1} + X_t X_t^\top$
        \For{$j = 1, 2, \ldots, m$}
            \State Sample $Z_t^j \sim \Pcal_R$
            \State Update $S_t^j = S_{t-1}^j + X_t ( Y_t +  Z_t^j)$\, and \,$\theta_t^j = V_t^{-1} S_t^j$
        \EndFor
    \EndFor
\end{algorithmic}
\end{algorithm}

\cref{alg:enssamp} describes linear ensemble sampling.
The learner maintains an ensemble of $m$ estimators, where each estimator fits perturbed rewards.
For the $j$-th estimator, a random vector $W^j \in \RR^d$ acts as an initial perturbation on the estimator, and a random variable $Z_t^j$ perturbs the reward at time $t$. 
Specifically, $\theta_t^j$ is the solution of the following minimization problem:
\begin{equation}
\label{eq:minimization problem}
    \mathop{\text{minimize}}_{\theta \in \RR^d} \lambda \left\| \theta -  W^j / \lambda \right\|_2^2 + \sum_{i=1}^{t} \left( X_i^\top \theta - \left( Y_i + Z_i^j \right) \right)^2
\end{equation}
When selecting an arm, one of the $m$ estimators is chosen according to an ensemble sampling distribution $\Jcal_t$ and acts greedily with respect to the sampled estimator.
Previous ensemble sampling algorithms~\citep{lu2017ensemble,qin2022analysis,janz2024ensemble} sample the estimators uniformly from the ensemble, but we allow any policy for selecting the estimator.
Further distinguishing from the algorithm presented in~\citet{janz2024ensemble}, we do not sample Rademacher random variables for symmetrization, making our algorithm simpler.
\\
Ensemble sampling is capable of being generalized to complex settings whenever solving minimization problem~\eqref{eq:minimization problem} is tractable.
Especially when incremental updates of the minimization problem are cheap, for instance, with neural networks or other gradient descent-based models, ensemble sampling can be an efficient exploration strategy.
The algorithm may simply store an ensemble of models, sample one to select an action, and then update the models incrementally based on the observed reward and generated perturbation.

\section{Regret Bound of Linear Ensemble Sampling}
\label{sec:regret bound of linear ensemble sampling}
Before we present the regret bound of linear ensemble sampling (Algorithm~\ref{alg:enssamp}), 
we present the following standard assumptions on the problem structure.
\begin{assumption}[Arm set and parameter]
\label{assm:arm set and parameter}
    $\Xcal$ is closed and for all $x \in \Xcal$, $\left\| x \right\|_2 \le 1$.
    There exists $S > 0$ such that
    $\left\| \theta^* \right\|_2 \le S$.
    Both bounds are known to the agent.
\end{assumption}
\begin{remark}
    Under Assumption~\ref{assm:arm set and parameter}, $\Xcal$ is a compact set.
    Therefore, we can define $x^* := \argmax_{x \in \Xcal} x^\top \theta^*$ and rewrite the definition of $R(T)$ as $\sum_{t=1}^T x^{*\top} \theta^* - X_t^\top \theta^*$.
\end{remark}

For a rigorous statement of the second assumption, we define several filtrations.
For $t \in [T] \cup \{0\}$, let $\Fcal_t^X := \sigma \left( X_1, \ldots X_t \right)$ and $\Fcal_t^\eta := \sigma \left( \eta_1, \ldots, \eta_t \right)$ be the $\sigma$-algebras generated by $X_i$ and $\eta_i$ up to time $t$ respectively.
We also define the $\sigma$-algebra generated by the algorithm's internal randomness up until the choice of $X_t$ as $\Fcal_t^A$.
Let $\Fcal_t := \sigma \left( \Fcal_t^A \cup \Fcal_t^X \cup \Fcal_t^\eta \right)$ be the $\sigma$-algebra generated by the first $t$ iterations of the interaction between the environment and the agent.
In addition, let $\Fcal_t^- :=  \sigma \left( \Fcal_t^A \cup \Fcal_t^X \cup \Fcal_{t-1}^\eta \right)$ be the $\sigma$-algebra generated in the same way as $\Fcal_t$, but excluding $\eta_t$.

\begin{assumption}[Noise]
    There exists $\sigma\ge 0$ such that $\eta_t$ is $\Fcal_{t}^-$-conditionally $\sigma$-subGaussian for all $t \in [T]$, i.e.,
    $\EE \left[ \exp ( s \eta_t) | \Fcal_{t}^-  \right] \le \exp \left( \sigma^2 s^2 / 2 \right)$
    holds almost surely for all $s \in \RR$.
\end{assumption}

Now, we define a value $\beta_t$ to describe the variance of the generated perturbation values.
Define a sequence $\{\beta_t(\delta) \}_{t=0}^\infty$ as $    \beta_t(\delta) := \sigma \sqrt{d \log \left( 1 + \frac{t}{d \lambda} \right) + 2 \log \frac{1}{\delta} } + \sqrt{\lambda} S$.
We may omit $\delta$ when its value is clear from the context.
The definition of $\beta_t$ comes from~\citet{abbasi2011improved} as a confidence radius of the ridge estimator, which we later specify in \cref{lma:theta hat concetration}.
We now present the regret bound of linear ensemble sampling (\cref{alg:enssamp}).

\begin{theorem}[Regret bound of linear ensemble sampling]
\label{thm:enssamp}
    Fix $\delta \in (0, 1]$. 
    Assume $|\Xcal| = K < \infty$ and run
    \cref{alg:enssamp} with $\lambda \ge 1$, $m \ge C (K \log T + \log \frac{1}{\delta} )$, $\Pcal_I = \Ncal(\boldsymbol{0}_d, \lambda \beta_T^2 I_d), \Pcal_R = \Ncal(0, \beta_T^2)$, and $\Jcal_t = \text{Unif}(m)$, where $C$ is a universal constant and $\text{Unif}(m)$ denotes the uniform distribution over $[m]$. 
    Then, with probability at least $1 - 4 \delta$, the cumulative regret of \cref{alg:enssamp} is 
    \begin{equation*}
        R(T) = \Ocal\!\left( (d \log T)^{\frac{3}{2}}\sqrt{T} \right).
    \end{equation*}
\end{theorem}

\paragraph{Discussion of Theorem~\ref{thm:enssamp}.} 

Theorem~\ref{thm:enssamp} shows that Algorithm~\ref{alg:enssamp} achieves a $\tilde{\Ocal}(d^{3/2}\sqrt{T})$ frequentist regret bound with an ensemble size of $m = \Omega(K \log T)$.
Importantly, our regret bound does not depend on $K$ or $m$ even logarithmically. 
Hence, this regret bound matches the state-of-the-art frequentist regret bound of linear Thompson sampling~\citep{agrawal2013thompson, abeille2017Linear}.
Our result is the first to establish
$\tilde{\Ocal}(d^{3/2}\sqrt{T})$ regret for linear ensemble sampling with an ensemble size sublinear in $T$, improving the previous bound by the factor $d$ compared to the existing result in \citet{janz2024ensemble}. We conjecture that $\tilde{\Ocal}(d^{3/2}\sqrt{T})$ regret is highly likely to be the best bound for linear ensemble sampling based on the negative result in \citet{hamidi2020frequentist} for LinTS.\footnote{\citet{hamidi2020frequentist} have shown that LinTS without the posterior variance inflation by $\sqrt{d}$ factor (compared to optimism-based algorithms such as LinUCB or OFUL \citep{abbasi2011improved}) can lead to a linear regret in $T$. That is, the frequentist regret bound of $\tilde{\Ocal}(d^{3/2}\sqrt{T})$ for LinTS is the best one can derive for the algorithm.}
Comparing with the algorithm in \citet{janz2024ensemble}, 
our version of linear ensemble sampling algorithm does not utilize Rademacher random variable for symmetrized perturbation.
This allows our algorithm to be simpler than that of \citet{janz2024ensemble}.
Partially due to this algorithmic difference, our regret analysis is quite distinct from the analysis of \citet{janz2024ensemble} (see the proof in Section~\ref{sec:proof of enssamp}).

As in~\citet{lu2017ensemble} and \citet{qin2022analysis}, we study the finite-armed problem setting.
Both studies analyze the excess regret of ensemble sampling compared to Thompson sampling through an information theoretical approach.
However, direct comparisons of the regret bounds are non-trivial.
The analysis by \citet{lu2017ensemble} includes an error admitted by the authors and~\citet{qin2022analysis} analyze the Bayesian regret, which is a weaker notion of regret than the frequentist regret that we analyze in this work.
Along with \citet{janz2024ensemble}, our result also makes progress in reducing the size of the ensemble compared to ~\citet{lu2017ensemble} and \citet{qin2022analysis}.
The size of the ensemble required by \citet{janz2024ensemble} is $\Theta(d \log T)$.
This requirement implies that their ensemble size may be smaller than ours when $K$ is larger than $d$ and also allows $K$ to be infinite.
However, it is important to note that their resulting regret bound of $\tilde{\Ocal}(d^{5/2}\sqrt{T})$ is clearly sub-optimal compared to regret bounds of other randomized exploration algorithms.
Theorem~\ref{thm:enssamp} achieves the tighter regret bound while simultaneously reducing the size of the ensemble.

\begin{table}[t!]
\label{table;enssamp comparison}
\caption{Comparison of regret bounds for linear ensemble sampling}
\begin{center}
\begin{threeparttable}
    \begin{tabular}{l l l l}
    \toprule
         Paper & Frequentist / Bayesian & Regret Bound & Ensemble Size  \\
         \midrule
         \citet{lu2017ensemble} & Frequentist & Invalid & Invalid
         \\
         \citet{qin2022analysis} & Bayesian & $\tilde{\Ocal}(\sqrt{d T \log K})$ & $\Omega( KT )$
         \\
         \citet{janz2024ensemble} & Frequentist & $\tilde{\Ocal}(d^{5/2} \sqrt{T})$ & $\Theta(d \log T)$
         \\
         \textbf{This work} & Frequentist & $\tilde{\Ocal}(d^{3/2} \sqrt{T})$ & $\Omega(K \log T)$
         \\
         \bottomrule
    \end{tabular}
\end{threeparttable}
\end{center}
\end{table}

\begin{remark}[\textbf{Counter-intuitive dependence on ensemble size} in \citet{janz2024ensemble}]\label{remark:janz_ensemble_size_increasing}
    The regret bound in  \citet{janz2024ensemble} actually grows super-linearly with the ensemble size, which is counter-intuitive.
    Their regret bound implies that as the ensemble size increases, the performance of the algorithm deteriorates.
    This fails to explain the superior empirical performance observed for ensemble sampling even with a large ensemble.
    On the contrary, our result in Theorem~\ref{thm:enssamp} does not show any performance degradation as the ensemble size increases.
\end{remark}

\begin{remark}[\textbf{Generalizability of perturbation distributions}]
\label{rmk:non gaussian}
    We show that Gaussian distribution for perturbation is not essential.
    The only properties of the Gaussian distribution we utilize are its tail probability and anti-concentration property, stated as \cref{lma:gaussian concentration lemma} and \cref{fact:gaussian} in \cref{sec:proof of enssamp}.
    Therefore, any other distributions exhibiting similar behaviors can  instead be adopted.
    In \cref{appx:beyond gaussian}, we rigorously demonstrate that any symmetric subGaussian distribution with lower-bounded variance can be employed, possibly at a cost of a constant factor.
    A large class of distributions, including uniform distribution, spherical distribution, Rademacher distribution, and centered binomial distribution with $p = 1/2$ satisfy this condition.
    This result can be of independent interest.
\end{remark}

\section{Analysis}
\label{sec:analysis}
\subsection{General Regret Analysis for Linear Bandits}

We begin by presenting a general regret bound for any algorithm that selects the best arm based on an estimated parameter. This result can be of independent interest.
This general bound and analysis serve as a general framework that includes the regret analysis of linear ensemble sampling (Theorem~\ref{thm:enssamp}).

\begin{theorem}[General regret bound for linear bandit algorithm]
\label{thm:general}
    Fix $T \in \NN$.
    Assume that at each time step $t \in [T]$, the agent chooses $X_t = \argmax_{x \in \Xcal} x^\top \theta_t$, where $\theta_t \in \RR^d$ is chosen by the agent under some (either deterministic or random) policy.
    Let $\lambda > 0$ and $V_t = \lambda I + \sum_{i=1}^t X_i X_i^\top$.
    Let $\left\{ \Ecal_{1, t} \right\}_{t=1}^T$ and $\left\{ \Ecal_{2, t} \right\}_{t=1}^T$ be sequences of events that satisfy two conditions:
    \begin{enumerate}
        \item (Concentration) There exists a constant $\gamma > 0$ such that 
        \begin{equation}
        \label{eq:Condition 1 concentration}
            \left\| \theta_t - \theta^* \right\|_{V_{t-1}} \ind \left\{ \Ecal_{1, t} \right\} \le \gamma
        \end{equation}
        holds almost surely for all $t \in [T]$.
        \item (Optimism) $\Ecal_{2, t} \in \Fcal_{t-1}$ holds and there exists a constant $p \in (0, 1]$ such that
        \begin{equation}
        \label{eq:Condition 2 Optimism}
        \PP \left( \left( x^{*\top}\theta^* \le X_t^\top \theta_t \text{ and } \Ecal_{1, t} \right) \text{ or }  \Ecal_{2, t}^{\Csf} \mid \Fcal_{t-1}  \right) \ge p
        \end{equation}
        holds almost surely for all $t \in [T]$.
    \end{enumerate}
    Take $\Ecal = \cap_{t=1}^T \left( \Ecal_{1, t} \cap \Ecal_{2, t} \right)$ and any $\delta \in (0, 1]$.
    Then, under the event $\Ecal$ and an additional event whose probability is at least $1 - \delta$, the cumulative regret is bounded as follows:
    \begin{equation}
        R(T) \le \gamma \left( 1 + \frac{2}{p} \right) \sqrt{ 2 d T \log \left( 1 + \frac{T}{d\lambda} \right) } + \frac{\gamma}{p} \sqrt{ \frac{2T}{\lambda} \log \frac{1}{\delta}}
        \, .
        \label{eq:regret bound}
    \end{equation}
\end{theorem}

\paragraph{Discussion of Theorem~\ref{thm:general}.}
The audience well-versed in the regret analysis of randomized algorithms such as TS and PHE would recognize that
bounding the regret using the probability of being optimistic is a standard procedure, also presented in Theorem~1 of~\citet{kveton2020perturbed} and Theorem~2 of~\citet{janz2024ensemble} as generalizations of the results in~\citet{agrawal2013thompson} and~\citet{abeille2017Linear} respectively.
However, our regret analysis offers a much more concise approach than the existing techniques, which can be of independent interest beyond the analysis of ensemble sampling.

\cref{thm:general} states that if the two conditions, specifically \textit{concentration} in~\eqref{eq:Condition 1 concentration} and \textit{optimism} in~\eqref{eq:Condition 2 Optimism}, are met, then the algorithm achieves $\sqrt{T}$ regret.
We provide the proof of \cref{thm:general} in \cref{appx:Proof of theorem 3}.
Our proof technique 
generalizes the well-studied analysis of 
\citet{abeille2017Linear}.
While their work poses conditions on a $d$-dimensional perturbation vector that is added to the ridge estimator, we do not assume the use of ridge regression or assume that the estimator is perturbed.
Instead, we only pose conditions on the final estimator the algorithm exploits.
Due to this generalization, \cref{thm:general} is even capable of inducing the regret bound of LinUCB~\citep{abbasi2011improved}, which always opts for an optimistic estimator, by setting $\gamma = 2 \beta_T$ and $p = 1$ with appropriate concentration events assigned to $\Ecal_{1, t}$ and $\Ecal_{2, t}$.
In addition, there are several improvements that simplify the proof which are worth noting.
To exploit the optimism condition, we apply Markov's inequality to a well-defined random variable.
This allows us to avoid using conditional expectations conditioned on the optimism event, which requires additional handling when the event has probability zero.
Furthermore, the theorem and its proof explicitly specify the high-probability event under which the analysis is conducted.
We also note that our proof does not require convex analysis studied in~\citet{abeille2017Linear}.

\begin{remark}[Role of event $\Ecal_{2, t}$]
    Previous results that utilize the probability of being optimistic~\citep{agrawal2013thompson,abeille2017Linear,kveton2020perturbed,janz2024ensemble} do not explicitly define events $\{\Ecal_{2, t}\}_t$.
    However, their existence is crucial in our analysis of linear ensemble sampling.
    Since the perturbation sequences are also part of the history in ensemble sampling, the probability of $\theta_t$ being optimistic may be extremely small under some events in $\Fcal_{t-1}$ that sample unfavorable sequences.
    The role of $\Ecal_{2, t}$ is to confine our analysis to the case where such undesirable events do not occur.
\end{remark}

\subsection{Proof of Regret Bound in Theorem~\ref{thm:enssamp}}
\label{sec:proof of enssamp}

We prove the regret bound of linear ensemble sampling stated in \cref{thm:enssamp}.
To apply \cref{thm:general}, high probabilities of the sequences of events, namely $\{\Ecal_{1, t}, \Ecal_{2, t}\}_{t=1}^T$, should be guaranteed with appropriate values of $\gamma$ and $p$.
We show that separate constraints can be imposed on the randomness of the rewards and the perturbations respectively to guarantee the probabilities of the events.
We begin by decomposing the estimator into two parts: one that fits the observed rewards and the other that perturbs the estimator.
\begin{align}
    \theta_t^j 
    &= V_t^{-1} S_t^j 
    = V_t^{-1} \left( W^j + \sum_{i=1}^t X_i \left( Y_i + Z_i^j \right) \right) 
     = V_t^{-1} \sum_{i=1}^t X_i Y_i + V_t^{-1} \left( W^j + \sum_{i=1}^t X_i Z_i^j \right) \notag
    \\
    & =: \hat{\theta}_t + \tilde{\theta}_t^j
    \, ,
    \label{eq:theta decomposition}
\end{align}
where we define $\hat{\theta}_t := V_t^{-1} \sum_{i=1}^t X_i Y_i$ and $\tilde{\theta}_t^j := V_t^{-1} ( W^j + \sum_{i=1}^t X_i Z_i^j )$.
$\hat{\theta}_t$ is the ridge regression estimator of the observed data, and its randomness mainly comes from the noise of the rewards, $\{\eta_i\}_{i=1}^t$.
$\tilde{\theta}_t^j$ is the perturbation added to $\hat{\theta}_t$, and its randomness comes from the generated perturbation, $W^j$ and $\{ Z_i^j\}_{i=1}^t$.
The following lemma states the well-known concentration result for the ridge estimator.

\begin{lemma}[Theorem~2 of~\citet{abbasi2011improved}]
\label{lma:theta hat concetration}
    Fix $\delta \in (0, 1]$.
    For $t \in \NN \cup \{0\}$, define a sequence of events
    with $\beta_t(\delta) = \sigma \sqrt{d \log \left( 1 + \frac{t}{d \lambda} \right) + 2 \log \frac{1}{\delta} } + \sqrt{\lambda} S$ as
    \begin{equation}
        \hat{\Ecal}_{t} := \left\{ \omega \in \Omega : \bigl\|  \hat{\theta}_t - \theta^* \bigr\|_{V_{t}} \le \beta_t(\delta) \right\} \notag
    \end{equation}
    and their intersection $\hat{\Ecal} := \bigcap_{t=0}^\infty \hat{\Ecal}_t$.
    Then, $\PP \big( \hat{\Ecal} \big) \ge 1 - \delta$.
\end{lemma}

Now, we address $\tilde{\theta}_t^j$.
Define a \textit{perturbation vector} that represents the perturbation sequence for each model as follows:
\begin{align}
    \Zb_t^j &:= \begin{pmatrix}
    \frac{1}{\sqrt{\lambda}}W^{j\top} & Z_1^j & \cdots & Z_{t-1}^j
    \end{pmatrix}^\top \in \RR^{d + t - 1} , \forall j \in [m]
    \label{eq:definition of Zb}
    \, .
\end{align}
Let $\Zb_t := \Zb_t^{j_t}$ so that $\Zb_t$ is the perturbation vector of the model chosen at time $t$.
The following lemma demonstrates that optimism condition~\eqref{eq:Condition 2 Optimism} can be satisfied by an anti-concentration property of $\Zb_t$ alone.

\begin{lemma}[Sufficient condition for optimism]
\label{lma:optimism lemma}
    For $t \in [T]$, define a vector $U_{t-1} \in \RR^{d + t - 1}$ by
    \begin{equation*}
        U_{t-1}^\top := x^{*\top} V_{t-1}^{-1} \begin{pmatrix}
            \sqrt{\lambda} I_d & X_1 & \cdots & X_{t-1}
        \end{pmatrix}
        \, .
    \end{equation*}
    Then, $x^{*\top} \theta^* \le X_t^{\top} \theta_t$ holds whenever there exists a constant $c > 0$ such that $U_{t-1}^\top \Zb_t \ge c \| U_{t-1} \|_2$ and $\| \theta^* - \hat{\theta}_{t-1} \|_{V_{t-1}} \le c$ hold.
\end{lemma}

We present a straightforward proof of \cref{lma:optimism lemma} in \cref{appx:Proof of optimism lemma}.
We significantly deviate from the analyses of~\citet{abeille2017Linear} and~\citet{janz2024ensemble} in the method of guaranteeing the optimism condition.
Their analyses require a $d$-dimensional perturbation vector to have a constant probability of having a positive component, so-called \textit{anti-concentrated}, in ``every'' possible direction in $\RR^d$ since they only prove the existence of a direction that implies optimism.
We observe and exploit the fact that it suffices to consider just ``one'' direction, specifically $U_{t-1}$, to produce an optimistic estimator.
Since $U_{t-1}$ depends only on the sequence of selected arms, the dependency between $U_{t-1}$ and $\Zb_t$ decouples when the dependency between $\{ X_t \}_{t=1}^T$ and $\{ \Zb_t \}_{t=1}^T$ is decoupled, which we later achieve by taking the union bound in a unique way.

The following lemma shows that concentration and anti-concentration properties of the perturbation are sufficient conditions for \cref{thm:general}.
We provide a sketch of its proof and defer the remaining details to \cref{appx:Proof of rpe lemma}.

\begin{lemma}
\label{lma:rpe result}
    Suppose the agent runs \cref{alg:enssamp} with some parameters.
    Fix $\tilde{\gamma} > 0$ and $p \in (0, 1]$.
    For each $t \in [T]$, define two events 
    \begin{align*}
        \tilde{\Ecal}_{1, t} & := \Bigl\{ \omega \in \Omega : \bigl\| \tilde{\theta}_{t-1}^{j_t} \bigr\|_{V_{t-1}} \le \tilde{\gamma} \Bigr\}
        \, ,
        \\
        \tilde{\Ecal}_{2, t} & := \left\{ \omega \in \Omega : \PP \left( U_{t-1}^\top {\Zb}_t \ge \beta_{t-1} \left\| U_{t-1}\right\|_2 \text{ and } \tilde{\Ecal}_{1, t} \mid \Fcal_{t-1}\right) \ge p \right\}
        \, ,
    \end{align*}
    where $U_{t-1}$ is defined as in \cref{lma:optimism lemma}.
    Take $\Ecal_{1, t} = \tilde{\Ecal}_{1, t} \cap \hat{\Ecal}_{t-1}$ and $\Ecal_{2, t} = \tilde{\Ecal}_{2, t} \cap \hat{\Ecal}_{t-1}$.
    Then, $\Ecal_{1, t}$ and $\Ecal_{2, t}$ satisfy concentration condition~\eqref{eq:Condition 1 concentration} with $\gamma = \tilde{\gamma} + \beta_T$ and optimism condition~\eqref{eq:Condition 2 Optimism} with the same value of $p$.
    Consequently, with probability at least $1 - 2 \delta - \PP (  \tilde{\Ecal}^\Csf)$, where $\tilde{\Ecal} := \cap_{t=1}^T (\tilde{\Ecal}_{1, t} \cap \tilde{\Ecal}_{2, t} )$, \cref{alg:enssamp} achieves regret bound~\eqref{eq:regret bound} of \cref{thm:general}.
\end{lemma}

\begin{remark}
    \cref{lma:rpe result} applies to any perturbation-based algorithm that exploits $\theta_t = \hat{\theta}_{t-1} + \tilde{\theta}_{t-1}$, where $\tilde{\theta}_{t-1}$ is a linear transform of a random perturbation vector $\Zb_t$.
    A version of linear Thompson sampling~\citep{abeille2017Linear} and LinPHE also fall into this category.
\end{remark}

\cref{lma:rpe result} shifts the problem of constructing the regret bound of \cref{alg:enssamp} to lower-bounding the probabilities of the two sequences of events, $\{\tilde{\Ecal}_{1, t}\}_{t=1}^T$ and $\{\tilde{\Ecal}_{2, t}\}_{t=1}^T$.
Note that $\tilde{\Ecal}_{1, t}$ and $\tilde{\Ecal}_{2, t}$ regard $\{X_i\}_{i=1}^{t-1}$ and $\Zb_t$ only, and are independent of further randomness of $\{ \eta_t\}_{t=1}^T$.

\begin{proof}[Sketch of Proof of \cref{lma:rpe result}]
    The concentration condition follows immediately by the triangle inequality.
    To show the optimism condition, we verify the following logical implication relationship:
    \begin{align*}
        \left( \left( U_{t-1}^\top {\Zb}_t \ge \beta_{t-1} \left\| U_{t-1} \right\|_2 \right) \land \tilde{\Ecal}_{1, t} \right) \lor \Ecal_{2, t}^\Csf
        \Rightarrow
        \left( \left( x^{*\top} \theta^* \le X_t^\top \theta_t \right) \land \Ecal_{1, t} \right) \lor \Ecal_{2, t}^\Csf
        \, ,
    \end{align*}
    where \cref{lma:optimism lemma} bridges the anti-concentration on the left-hand side to the optimism on the right-hand side.
    This implication relationship is converted to the following probability inequality:
    \begin{align*}
        \PP \left( \left( ( x^{*\top} \theta^* \le X_t^\top \theta_t ) \land \Ecal_{1, t} \right) \lor \Ecal_{2, t}^\Csf \mid \Fcal_{t-1} \right) & \ge 
        \PP \left(\left( ( U_{t-1}^\top {\Zb}_t \ge \beta_{t-1} \left\| U_{t-1} \right\|_2 ) \land \tilde{\Ecal}_{1, t} \right) \lor \Ecal_{2, t}^\Csf \mid \Fcal_{t-1} \right)
        \, .
    \end{align*}
    By the definition of $\tilde{\Ecal}_{2, t}$, the right-hand side is bounded below by $p$, implying optimism condition~\eqref{eq:Condition 2 Optimism}.
    The probability of failure is bounded by the union bound and \cref{lma:theta hat concetration}.
\end{proof}

\cref{thm:enssamp} employs the Gaussian perturbation for a concrete instantiation of \cref{alg:enssamp}.
We define two values $\tilde{\gamma}_T$ and $\gamma_T$, which serve as the confidence radii of $\tilde{\theta}_t$ and $\theta_t$ for $t \in [T]$ under the Gaussian perturbation.
\begin{align}
    \tilde{\gamma}_T := \beta_T \left( \sqrt{ d \log \left( 1 + \frac{T}{d \lambda} \right) + 2 \log \frac{2T}{\delta}} + \sqrt{d} + \sqrt{2 \log \frac{2T }{\delta}} \right), && \gamma_T := \tilde{\gamma}_T + \beta_T
    \label{eq:definition of gamma_T}
    \, .
\end{align}
Note that in terms of $d$ and $T$, both $\tilde{\gamma}_T$ and $\gamma_T$ are in $\Ocal\left( d \log T \right)$.
\cref{lma:gaussian concentration lemma} illustrates the concentration result.
Its proof is a simple application of \cref{lma:theta hat concetration}, and is presented in \cref{appx:Proof of gaussian concentration}.

\begin{lemma}
\label{lma:gaussian concentration lemma}
    Suppose \cref{alg:enssamp} is run with parameters specified in \cref{thm:enssamp}.
    Fix $t \in [T]$ and $j \in [m]$.
    Suppose the sequence of arms $X_1, \ldots, X_{t}$ is chosen arbitrarily randomly, not necessarily by the agent.
    Let $\tilde{\theta}_t^j$ be defined as in Eq.~\eqref{eq:theta decomposition}.
    Then, with probability at least $1 - \delta/ T$, $\| \tilde{\theta}_{t}^j \|_{V_{t}} \le \tilde{\gamma}_T$ holds.
\end{lemma}

The following fact describes an anti-concentration property of the Gaussian distribution, which follows from the fact that a linear combination of independent Gaussians is again Gaussian.

\begin{fact}
\label{fact:gaussian}
    If $Z \sim \Ncal(0, \alpha^2 I_n)$ for some $\alpha \ge 0$ and $u \in \RR^n$ is a fixed vector for some $n \in \NN$, then $\PP \left(u^\top Z \ge \alpha \left\| u \right\|_2 \right) \ge \PP \left( z \ge 1\right) =: p_N$, where $z \sim \Ncal(0, 1)$.
    We note that $p_N \ge 0.15$.
\end{fact}

All the building blocks we need to prove \cref{thm:enssamp} are ready.
The proof illustrates that the events specified in \cref{lma:rpe result} occur with high probability.

\begin{proof}[Proof of \cref{thm:enssamp}]
    Define $\tilde{\Ecal}_{1, t}$ and $\tilde{\Ecal}_{2, t}$ as in \cref{lma:rpe result} with $\tilde{\gamma} = \tilde{\gamma}_T$ and $p = p_N/4$, where $\tilde{\gamma}_T$ is defined in Eq.~\eqref{eq:definition of gamma_T} and $p_N$ is defined in \cref{fact:gaussian}.
    We show that these events occur with high probability, and the rest follows from \cref{lma:rpe result}.
    For the sake of the analysis, assume that $\delta / T \le p_N / 2 \approx 0.08$, which holds whenever $T \ge 14$ or $\delta < 0.07$.
    \\
    Assume that the perturbation values $W^j$ and $Z_t^j$ are $\Fcal_0^A$-measurable for all $j \in [m]$ and $t \in [T]$.
    An interpretation of this assumption is that the algorithm samples all the required values in advance.
    Note that we still obtain an equivalent algorithm and this modification need not actually take place in the execution.
    Under this assumption, the uniform sampling of $j_t \sim \Jcal_t$ is the only source of randomness regarding the choice of $\theta_t$ and $X_t$ when conditioned on the history $\Fcal_{t-1}$.
    It may seem unintuitive, but this modification simplifies the proof because we only need to deal with $j_t$.
    \\
    We first lower-bound the probability of $\tilde{\Ecal}_{1, t}$.
    When $j \in [m]$ is fixed, we can apply \cref{lma:gaussian concentration lemma} and obtain $\PP ( \| \tilde{\theta}_{t-1}^j \|_{V_{t-1}} \le \tilde{\gamma}_T ) \ge 1 - \delta / T$.
    Since $j_t$ is sampled independently of $\tilde{\theta}_{t-1}^j$, it holds that
    \begin{align}
        \PP \left( \tilde{\Ecal}_{1, t} \right) = \sum_{j=1}^m \PP \left( j_t = j, \bigl\| \tilde{\theta}_{t-1}^j \bigr\|_{V_{t-1}} \le \tilde{\gamma}_T \right) = \sum_{j=1}^m \frac{1}{m} \PP \left( \bigl\| \tilde{\theta}_{t-1}^j \bigr\|_{V_{t-1}} \le \tilde{\gamma}_T \right) \ge 1 - \frac{\delta}{T}
        \, .
        \label{eq:Ecal_1 bound}
    \end{align}
    Now, we bound the probability of $\tilde{\Ecal}_{2, t}$.
    Fix $j \in [m]$.
    Recall that $\Zb_t^j$ is the perturbation vector of the $j$-th model, defined in Eq.~\eqref{eq:definition of Zb}.
    The choice of Gaussian perturbation implies that $\Zb_t^j \sim \Ncal(\boldsymbol{0}_{d + t - 1}, \beta_T^2 I_{d + t -1})$.
    Suppose that the sequence of arms $X_1, \ldots, X_T$ is fixed.
    Then, we can apply \cref{fact:gaussian}, obtaining that $\PP( U_{t-1}^\top {\Zb}_t^j \ge \beta_{t-1} \| U_{t-1}\|_2 ) \ge p_N$, where the probability is measured over the randomness of the perturbation sequence ${\Zb}_t^j$.
    Let $I^j_t := \ind \{ ( U_{t-1}^\top {\Zb}_t^j \ge \beta_{t-1} \| U_{t-1}\|_2) \land  ( \bigl\| \tilde{\theta}_{t-1}^{j} \bigr\|_{V_{t-1}} \le \tilde{\gamma}_T ) \} $.
    Then, we have that
    \begin{align}
        \PP \bigl( I^j_t = 1 \bigr) \ge \PP \bigl(U_{t-1}^\top {\Zb}_t^j \ge \beta_{t-1} \| U_{t-1}\|_2 \bigr) - \PP \Bigl( \bigl\| \tilde{\theta}_{t-1}^j \bigr\|_{V_{t-1}} > \tilde{\gamma}_T \Bigr) \ge p_N - \delta / T \ge p_N / 2
        \, ,
        \label{eq:optimism condition bound}
    \end{align}
    where the first inequality uses that $\PP(A \cap B) \ge \PP(A) - \PP(B^\Csf)$ holds for any events $A$ and $B$, and the last inequality holds by the assumption $\delta / T \le p_N / 2$.
    However, as we assumed that the perturbation sequence is $\Fcal_0^A$-measurable, it is $\Fcal_{t-1}$-measurable, hence $I^j_t$ is also $\Fcal_{t-1}$-measurable.
    It means that the value of $I^j_t$ is determined when the history up to time $t-1$ is fixed.
    The only remaining source of randomness in choosing $\theta_t$ conditioned on $\Fcal_{t-1}$ is the sampling of $j_t \sim \Jcal_t$.
    Therefore, it holds that
    \begin{align}
    \label{eq:conditional probability equality 2}
        \PP \left( \left( U_{t-1}^\top {\Zb}_t^{j_t} \ge \beta_{t-1} \| U_{t-1} \|_2 \right) \land \left( \bigl\| \tilde{\theta}_{t-1}^{j_t} \bigr\|_{V_{t-1}} \le \tilde{\gamma}_T \right) \mid \Fcal_{t-1} \right) = \frac{1}{m} \sum_{j=1}^m I^j_t
        \, .
    \end{align}
    Since we have verified that the expectation of the right-hand side is greater than $p_N / 2$, the Azuma-Hoeffding inequality (\cref{lma:Azuma-Hoeffding inequality}) implies that $\PP ( \frac{1}{m} \sum_{j=1}^m I^j_t < p_N / 4 ) \le \exp \left( - p_N^2 m / 8 \right)$.
    Recall that this result is obtained assuming that $X_1, \ldots, X_T$ are fixed.
    We take the union bound over all possible sequences of arms.
    However, a na\"ive union bound multiplies $K^T$ to the failure probability, which leads to an undesirable result of $m$ scaling linearly with $T$.
    We present the following proposition inspired by an observation from~\citet{lu2017ensemble} that a permutation of selected arms can be regarded as equivalent.
    We note that the strong result of Lemma~6 in~\citet{lu2017ensemble} is not applicable to our setting since we do not assume that $\{\eta_t\}_{t=1}^T$ is distributed identically nor independently.
    Although we present all the main ideas to support \cref{prop:optimism} in this section, there may be a few points that readers find require further justification.
    Due to limited space, we provide a full, rigorous justification of \cref{prop:optimism} in \cref{appx:rigorous explanation}, where we present a different perspective on the sampling of perturbation.
    \begin{proposition}
    \label{prop:optimism}
        There exists an event $\Ecal_2^*$ such that under $\Ecal_2^*$, $\frac{1}{m} \sum_{j=1}^m I_t^j \ge p_N / 4$ holds for $t = 1, \ldots, T$ and $\PP ( \Ecal_2^{*\Csf}) \le T^K \exp ( - p_N^2 m / 8 )$.
    \end{proposition}
    The key observation in \cref{prop:optimism} is that the perturbation vector consists of i.i.d. components, hence its distribution is invariant under independent permutations.
    Therefore, the distributions of $\tilde{\theta}_{t-1}^j$ and $U_{t-1}^\top {\Zb}_t^j$ remain invariant under the permutation of selected arms.
    Although the sequence of arms and the perturbation vector are not independent as a whole, the permutation that sorts the selected arms preserves the distribution of $\Zb_t$ since $X_t$ and $Z_t$ are independent for all $t \in [T]$.
    The number of equivalence classes up to permutation over sequences of arms with lengths at most $T-1$ is less than $T^K$, since each arm can be selected $0$ to $T-1$ times inclusively.
    Therefore, we take the union bound over the $T^K$ sequences of arms and attain $\Ecal_2^*$.
    \\
    Taking $m = \frac{8}{p_N^2} ( K \log T + \log \frac{1}{\delta} )$, we obtain that $\PP( \Ecal_2^{*\Csf}) \le \delta$. 
    Eq.~\eqref{eq:conditional probability equality 2} implies that $\PP ( \cap_{t=1}^T \tilde{\Ecal}_{2, t} ) \ge \PP ( \Ecal_2^* ) \ge 1 - \delta$.
    Therefore, by \cref{lma:rpe result}, with probability at least $1 - 4 \delta$, the cumulative regret is bounded as follows:
    \begin{align*}
        R(T) \le  \gamma_T \left( 1 + \frac{8}{p_N} \right) \sqrt{ 2 d T \log \left( 1 + \frac{T}{d\lambda} \right) } + \frac{4 \gamma_T}{p_N} \sqrt{ \frac{2T}{\lambda} \log \frac{1}{\delta}} 
        = O\left((d \log T) ^{\frac{3}{2}} \sqrt{T} \right)
        \, .
    \end{align*}
\end{proof}

\section{Ensemble Sampling and Perturbed-History Exploration}
\label{sec:reduction to phe}
\begin{algorithm}[t!]
\caption{Linear Perturbed-History Exploration (LinPHE)}
\label{alg:phe}
\begin{algorithmic}
    \State \textbf{Input} : regularization parameter $\lambda > 0$, initial perturbation distribution $\Pcal_I$ on $\RR^d$,
    \State \qquad reward-perturbation distribution $\Pcal_R$ on $\RR$
    \State Initialize $V_0 = \lambda I_d$
    \For{$t  = 1, 2, \ldots T$}
        \State Sample $W_t \sim \Pcal_I$, $Z_{t, 1}, \ldots, Z_{t, t-1} \stackrel{\iid}{\sim} \Pcal_R$
        \State Update $\theta_t = V_{t-1}^{-1} \left( W_t + \sum_{i=1}^{t-1} X_i (Y_i + Z_{t, i}) \right)$
        \State Pull arm $X_t = \argmax_{x \in \Xcal} x^\top \theta_t$, and observe $Y_t$
        \State Update $V_t = V_{t-1} + X_t X_t^\top$
    \EndFor
\end{algorithmic}
\end{algorithm}

In this section, we rigorously investigate the relationship between linear ensemble sampling and LinPHE.
A generalized version of LinPHE is described in \cref{alg:phe}.
Note that the perturbed estimator, $\theta_t = V_{t-1}^{-1} \big( W_t + \sum_{i=1}^{t-1} X_i (Y_i + Z_{t, i}) \big)$, resembles the estimator of linear ensemble sampling, which becomes evident when compared with Eq.~\eqref{eq:theta decomposition}.
The main difference is that in LinPHE (\cref{alg:phe}), the perturbation sequence is generated independently of the history at every time step, whereas in linear ensemble sampling (\cref{alg:enssamp}), the sequence is not renewed but is incremented at each time step.
However, we further observe that in linear ensemble sampling, as long as an estimator is not sampled for the arm selection, its perturbation sequence is independent of the selected arms and rewards.
This implies that the estimator in the ensemble that is selected for the first time is equivalent to the estimator computed by the policy of LinPHE.
Specifically, if the $j$-th estimator is selected for the first time at time step $t$, then the perturbation values of the estimator, $W^j$ and $\{Z_i^j\}_{i=1}^{t-1}$, have had no effect on previous interactions.
Therefore, newly sampling them as $W_t^j \sim \Pcal_I$ and $\{Z_{t, i}^j\}_{i=1}^{t-1} \stackrel{\iid}{\sim} \Pcal_R$, as in LinPHE, does not alter future interactions.
We conclude that in the case where the ensemble size is greater than or equal to $T$, linear ensemble sampling becomes equivalent to LinPHE by selecting the estimators in a round robin.

\begin{proposition}
\label{prop:enssamp is phe}
    Linear ensemble sampling (\cref{alg:enssamp}) with $m = T$ and deterministic policy of choosing a model, e.g., $\Jcal_t \equiv t$ for $t = 1, \ldots, T$, is equivalent to LinPHE (\cref{alg:phe}).
\end{proposition}

\cref{prop:enssamp is phe} shows that LinPHE is a special case of linear ensemble sampling and provides insightful consequences in both directions of the equivalence.
To our best knowledge, \cref{prop:enssamp is phe} is the first result to formally demonstrate the relationship between linear ensemble sampling and LinPHE.

\textbf{Linear ensemble sampling with $T$ models is LinPHE: }
Since an ensemble of $T$ models is equivalent to LinPHE which achieves a regret bound $\tilde{\Ocal}(d\sqrt{T \log K})$, the ensemble size larger than $T$ is not necessary.
This implication certainly emphasizes the sub-optimal requirements of the ensemble size in ~\citet{lu2017ensemble,qin2022analysis}.
Even when $K > T$ in our problem setting, this equivalence provides the ground for upper bounding the ensemble size by $T$.

\textbf{LinPHE is linear ensemble sampling with $T$ models: }
Conversely, since LinPHE can be regarded as linear ensemble sampling with $T$ models, it is possible to derive a regret bound of LinPHE by following the proof of \cref{thm:enssamp}.
We present \cref{cor:phe}, which states a regret bound of $\tilde{\Ocal}( d^{3/2} \sqrt{T})$ for LinPHE.
Note that in this case, the regret bound is independent of $K$.

\begin{corollary}[Regret bound of LinPHE]
\label{cor:phe}
    Fix $\delta \in (0, 1]$.
     \cref{alg:phe} with $\lambda \ge 1$, $\Pcal_I = \Ncal(\boldsymbol{0}_d, \lambda \beta_T^2 I_d)$ and $\Pcal_R = \Ncal(0, \beta_T^2)$ achieves $\Ocal((d \log T)^{3/2}\sqrt{T})$ cumulative regret with probability at least $1 - 3 \delta$.
\end{corollary}

\paragraph{Discussion of \cref{cor:phe}.}
\citet{kveton2020perturbed} provide a $\tilde{\Ocal}(d \sqrt{T \log K})$ regret bound when the number of arms is finite.
We prove that LinPHE achieves a $\tilde{\Ocal} ( d^{3/2} \sqrt{T} )$ regret bound independent of the number of arms, using our analysis framework. 
It is widely observed that assuming the size of the arm set to be $K$ may lead to an interchanging of a $\sqrt{d}$ factor with a $\sqrt{\log K}$ factor in the regret bound~\citep{agrawal2013thompson}, although attaining such reduction may not always be done in a trivial manner~\citep{auer2002using,chu2011contextual,bubeck2012towards}. It it worth noting that 
$\tilde{\Ocal} ( d^{3/2} \sqrt{T} )$ regret bound was previously achieved for LinPHE~\citep{janz2024exploration}.
Our focus is not merely on proving another regret bound for LinPHE, but rather on highlighting the close relationship between linear ensemble sampling and LinPHE, which, to our knowledge, has been overlooked in the literature.
\\
The proof of \cref{cor:phe} follows the proof of \cref{thm:enssamp}.
Note that the latter part of the proof of \cref{thm:enssamp} focuses on decoupling the dependency between $\{X_t\}_{t=1}^T$ and $\Zb_t$.
However, in the case of LinPHE, they are already independent since the perturbation sequence is freshly sampled at every time step, enabling a more elegant and concise proof.
Especially, as it skips the parts that require the number of arms to be finite, for instance the use of \cref{prop:optimism}, \cref{cor:phe} holds even when the number of arms is infinite.
The whole proof is presented in \cref{sec:proof of phe}.

\section{Conclusion}
\label{appx:conclusion}

We prove that linear ensemble sampling achieves a $\tilde{\Ocal}(d^{3/2} \sqrt{T})$ regret bound, marking the first such result in the frequentist setting and matching the best-known regret bound for randomized algorithms. 
The required ensemble size scales logarithmically with the time horizon as $\Omega(K \log T)$. 
Additionally, we expand our analysis to LinPHE, demonstrating that it is a special case of linear ensemble sampling with an ensemble of \( T \) models, achieving the same regret bound of $\tilde{\Ocal}(d^{3/2} \sqrt{T})$.
While our work focuses on linear bandits, ensemble sampling applications have shown superior performance in more complex settings. This suggests that theoretical extensions beyond the linear setting are worth pursuing, with our results providing an important foundation for possibly understanding these extensions. 
Extending the results to general contextual settings, where the arm set may change over time with potentially non-linear reward functions, represents a promising direction for future work.


\section*{Acknowledgements}
This work was supported by the National Research Foundation of Korea(NRF) grant funded by the Korea government(MSIT) (No. 2022R1C1C1006859, 2022R1A4A1030579, and RS-2023-00222663) and by AI-Bio Research Grant through Seoul National University.

\bibliography{references.bib}

\begin{thebibliography}{27}
\providecommand{\natexlab}[1]{#1}
\providecommand{\url}[1]{\texttt{#1}}
\expandafter\ifx\csname urlstyle\endcsname\relax
  \providecommand{\doi}[1]{doi: #1}\else
  \providecommand{\doi}{doi: \begingroup \urlstyle{rm}\Url}\fi

\bibitem[Abbasi-Yadkori et~al.(2011)Abbasi-Yadkori, P{\'a}l, and Szepesv{\'a}ri]{abbasi2011improved}
Yasin Abbasi-Yadkori, D{\'a}vid P{\'a}l, and Csaba Szepesv{\'a}ri.
\newblock Improved algorithms for linear stochastic bandits.
\newblock \emph{Advances in Neural Information Processing Systems}, 24:\penalty0 2312--2320, 2011.

\bibitem[Abeille and Lazaric(2017)]{abeille2017Linear}
Marc Abeille and Alessandro Lazaric.
\newblock {Linear Thompson Sampling Revisited}.
\newblock In \emph{Proceedings of the 20th International Conference on Artificial Intelligence and Statistics}, volume~54, pages 176--184. PMLR, PMLR, 2017.

\bibitem[Agrawal and Goyal(2012)]{agrawal2012analysis}
Shipra Agrawal and Navin Goyal.
\newblock Analysis of thompson sampling for the multi-armed bandit problem.
\newblock In \emph{Conference on learning theory}, pages 39--1. JMLR Workshop and Conference Proceedings, 2012.

\bibitem[Agrawal and Goyal(2013)]{agrawal2013thompson}
Shipra Agrawal and Navin Goyal.
\newblock Thompson sampling for contextual bandits with linear payoffs.
\newblock In \emph{International conference on machine learning}, pages 127--135. PMLR, 2013.

\bibitem[Auer(2002)]{auer2002using}
Peter Auer.
\newblock Using confidence bounds for exploitation-exploration trade-offs.
\newblock \emph{Journal of Machine Learning Research}, 3\penalty0 (Nov):\penalty0 397--422, 2002.

\bibitem[Bubeck et~al.(2012)Bubeck, Cesa-Bianchi, and Kakade]{bubeck2012towards}
S{\'e}bastien Bubeck, Nicolo Cesa-Bianchi, and Sham~M Kakade.
\newblock Towards minimax policies for online linear optimization with bandit feedback.
\newblock In \emph{Conference on Learning Theory}, pages 41--1. JMLR Workshop and Conference Proceedings, 2012.

\bibitem[Chapelle and Li(2011)]{chapelle2011empirical}
Olivier Chapelle and Lihong Li.
\newblock An empirical evaluation of thompson sampling.
\newblock \emph{Advances in neural information processing systems}, 24, 2011.

\bibitem[Chu et~al.(2011)Chu, Li, Reyzin, and Schapire]{chu2011contextual}
Wei Chu, Lihong Li, Lev Reyzin, and Robert Schapire.
\newblock Contextual bandits with linear payoff functions.
\newblock In \emph{Proceedings of the Fourteenth International Conference on Artificial Intelligence and Statistics}, pages 208--214. JMLR Workshop and Conference Proceedings, 2011.

\bibitem[Hamidi and Bayati(2020)]{hamidi2020frequentist}
Nima Hamidi and Mohsen Bayati.
\newblock On frequentist regret of linear thompson sampling.
\newblock \emph{arXiv preprint arXiv:2006.06790}, 2020.

\bibitem[Janz et~al.(2024)Janz, Litvak, and Szepesv{\'a}ri]{janz2024ensemble}
David Janz, Alexander Litvak, and Csaba Szepesv{\'a}ri.
\newblock Ensemble sampling for linear bandits: small ensembles suffice.
\newblock \emph{Advances in Neural Information Processing Systems}, 37:\penalty0 23679--23704, 2024.

\bibitem[Kveton et~al.(2019)Kveton, Szepesvari, Ghavamzadeh, and Boutilier]{kveton2019perturbed}
Branislav Kveton, Csaba Szepesvari, Mohammad Ghavamzadeh, and Craig Boutilier.
\newblock Perturbed-history exploration in stochastic multi-armed bandits.
\newblock In \emph{International Joint Conference on Artificial Intelligence}, 2019.
\newblock URL \url{https://api.semanticscholar.org/CorpusID:67856126}.

\bibitem[Kveton et~al.(2020{\natexlab{a}})Kveton, Szepesv{\'a}ri, Ghavamzadeh, and Boutilier]{kveton2020perturbed}
Branislav Kveton, Csaba Szepesv{\'a}ri, Mohammad Ghavamzadeh, and Craig Boutilier.
\newblock Perturbed-history exploration in stochastic linear bandits.
\newblock In \emph{Uncertainty in Artificial Intelligence}, pages 530--540. PMLR, 2020{\natexlab{a}}.

\bibitem[Kveton et~al.(2020{\natexlab{b}})Kveton, Zaheer, Szepesvari, Li, Ghavamzadeh, and Boutilier]{kveton2020randomized}
Branislav Kveton, Manzil Zaheer, Csaba Szepesvari, Lihong Li, Mohammad Ghavamzadeh, and Craig Boutilier.
\newblock Randomized exploration in generalized linear bandits.
\newblock In \emph{International Conference on Artificial Intelligence and Statistics}, pages 2066--2076. PMLR, 2020{\natexlab{b}}.

\bibitem[Lattimore and Szepesv{\'a}ri(2020)]{lattimore2020bandit}
Tor Lattimore and Csaba Szepesv{\'a}ri.
\newblock \emph{Bandit algorithms}.
\newblock Cambridge University Press, 2020.

\bibitem[Laurent and Massart(2000)]{laurent2000adaptive}
Beatrice Laurent and Pascal Massart.
\newblock Adaptive estimation of a quadratic functional by model selection.
\newblock \emph{Annals of statistics}, pages 1302--1338, 2000.

\bibitem[Lu and Van~Roy(2017)]{lu2017ensemble}
Xiuyuan Lu and Benjamin Van~Roy.
\newblock Ensemble sampling.
\newblock \emph{Advances in Neural Information Processing Systems}, 30, 2017.

\bibitem[Lu et~al.(2018)Lu, Wen, and Kveton]{lu2018efficient}
Xiuyuan Lu, Zheng Wen, and Branislav Kveton.
\newblock Efficient online recommendation via low-rank ensemble sampling.
\newblock In \emph{Proceedings of the 12th ACM Conference on Recommender Systems}, pages 460--464, 2018.

\bibitem[Osband et~al.(2016)Osband, Blundell, Pritzel, and Van~Roy]{osband2016deep}
Ian Osband, Charles Blundell, Alexander Pritzel, and Benjamin Van~Roy.
\newblock Deep exploration via bootstrapped dqn.
\newblock \emph{Advances in Neural Information Processing Systems}, 29, 2016.

\bibitem[Osband et~al.(2018)Osband, Aslanides, and Cassirer]{osband2018randomized}
Ian Osband, John Aslanides, and Albin Cassirer.
\newblock Randomized prior functions for deep reinforcement learning.
\newblock \emph{Advances in Neural Information Processing Systems}, 31, 2018.

\bibitem[Osband et~al.(2019)Osband, Van~Roy, Russo, Wen, et~al.]{osband2019deep}
Ian Osband, Benjamin Van~Roy, Daniel~J Russo, Zheng Wen, et~al.
\newblock Deep exploration via randomized value functions.
\newblock \emph{Journal of Machine Learning Research}, 20\penalty0 (124):\penalty0 1--62, 2019.

\bibitem[Qin et~al.(2022)Qin, Wen, Lu, and Van~Roy]{qin2022analysis}
Chao Qin, Zheng Wen, Xiuyuan Lu, and Benjamin Van~Roy.
\newblock An analysis of ensemble sampling.
\newblock \emph{Advances in Neural Information Processing Systems}, 35:\penalty0 21602--21614, 2022.

\bibitem[Russo and Van~Roy(2014)]{russo2014learning}
Daniel Russo and Benjamin Van~Roy.
\newblock Learning to optimize via posterior sampling.
\newblock \emph{Mathematics of Operations Research}, 39\penalty0 (4):\penalty0 1221--1243, 2014.

\bibitem[Russo et~al.(2018)Russo, Van~Roy, Kazerouni, Osband, Wen, et~al.]{russo2018tutorial}
Daniel~J Russo, Benjamin Van~Roy, Abbas Kazerouni, Ian Osband, Zheng Wen, et~al.
\newblock A tutorial on thompson sampling.
\newblock \emph{Foundations and Trends{\textregistered} in Machine Learning}, 11\penalty0 (1):\penalty0 1--96, 2018.

\bibitem[Thompson(1933)]{thompson1933likelihood}
William~R Thompson.
\newblock On the likelihood that one unknown probability exceeds another in view of the evidence of two samples.
\newblock \emph{Biometrika}, 25\penalty0 (3/4):\penalty0 285--294, 1933.

\bibitem[Wan et~al.(2023)Wan, Wei, Kveton, and Song]{wan2023multiplier}
Runzhe Wan, Haoyu Wei, Branislav Kveton, and Rui Song.
\newblock Multiplier bootstrap-based exploration.
\newblock In \emph{International Conference on Machine Learning}, pages 35444--35490. PMLR, 2023.

\bibitem[Zhou et~al.(2024)Zhou, Hao, Wen, Zhang, and Sun]{zhou2024stochastic}
Jie Zhou, Botao Hao, Zheng Wen, Jingfei Zhang, and Will~Wei Sun.
\newblock Stochastic low-rank tensor bandits for multi-dimensional online decision making.
\newblock \emph{Journal of the American Statistical Association}, pages 1--14, 2024.

\bibitem[Zhu and Van~Roy(2023)]{zhu2023deep}
Zheqing Zhu and Benjamin Van~Roy.
\newblock Deep exploration for recommendation systems.
\newblock In \emph{Proceedings of the 17th ACM Conference on Recommender Systems}, pages 963--970, 2023.

\end{thebibliography}
\bibliographystyle{plainnat}

\newpage

\part*{Appendix}

\appendix

\section{Notations}
We summarize the notations in this paper in \cref{table:specific notations} and \cref{table:generic notations}.

\begin{table*}[ht]
    \centering
    \caption{Notations specific to this paper}
    \label{table:specific notations}
    \begin{tabular}{ c l }
        \toprule
        Linear Bandit
        \\
        \hline
        $\Xcal$ & Set of arms
        \\
        $\theta^*$ & True parameter vector
        \\ 
        $x^*$ & Optimal arm
        \\
        $X_t$ & Chosen arm at time $t$
        \\
        $Y_t$ & Observed reward at time $t$
        \\
        $\eta_t$ & Zero-mean noise at time $t$
        \\
        $\sigma$ & SubGaussian variance proxy of $\eta_t$
        \\
        $d$ & Dimension of arms and true parameter vector
        \\
        $K$ & Number of arms (if $|\Xcal| < \infty$)
        \\
        $T$ & Time horizon 
        \\
        $R(T) $ & Cumulative regret
        \\
        \hline
        Algorithm
        \\
        \hline
        $\lambda$ & Regularization parameter
        \\
        $V_t$ & $\lambda I_d + \sum_{i=1}^t X_i X_i^\top$
        \\
        $\theta_t$ & Perturbed estimator
        \\
        $W^j$ / $W_t$ & Initial perturbation vector
        \\
        $Z_t^j$ / $Z_{t, i}$ & Reward perturbation
        \\
        $\Pcal_I$ & Distribution for initial perturbation $W$
        \\
        $\Pcal_R$ & Distribution for reward perturbation $Z_t$
        \\
        $\Jcal_t$ & Sampling policy at time $t$
        \\
        \hline
        Analysis
        \\
        \hline
        $\delta$ & Probability of failure
        \\
        $\Fcal_t^X$ & $\sigma(\{X_i\}_{i=1}^t)$
        \\
        $\Fcal_t^\eta$ & $\sigma(\{\eta_i\}_{i=1}^t)$
        \\
        $\Fcal_t^A$ & $\sigma$-algebra generated by algorithm's internal randomness until time $t$
        \\
        $\Fcal_t$ & $\sigma(\Fcal_t^A \cup \Fcal_t^X \cup \Fcal_t^\eta )$, $\sigma$-algebra generated by the first $t$-iterations of interaction
        \\
        $\hat{\theta}_t$ & Ridge regression estimator by $t$ samples
        \\
        $\tilde{\theta}_t$ & Perturbation in the estimator, $\theta_{t+1} - \hat{\theta}_t$
        \\
        $\beta_t$ & Confidence radius of $\hat{\theta}_t$
        \\
        $\gamma$ & Confidence radius of $\theta_t$
        \\
        $\tilde{\gamma}$ & Confidence radius of $\tilde{\theta}_t$
        \\
        $p$ & Probability of optimistic arm selection
        \\
        $p_N$ & $\PP( Z \ge 1) \approx 0.15$ where $Z \sim \Ncal(0, 1)$
        \\
        $\Ecal$ & High-probability event
        \\
        $\Ecal_{1, t}$ & Concentration event
        \\
        $\Ecal_{2, t}$ & Optimism event
        \\
        $\hat{\Ecal}_t$ & Concentration event of $\hat{\theta}_t$
        \\
        $\tilde{\Ecal}_{1, t}$ & Concentration event of $\tilde{\theta}_{t-1}$
        \\
        $\tilde{\Ecal}_{2, t}$ & Anti-concentration event of $\tilde{\theta}_{t-1}$
        \\
        $\Zb_t$ & Perturbation vector at time $t$
        \\
        $\Phi_t$ & $\begin{pmatrix}
            \sqrt{\lambda} I_d & X_1 & \cdots & X_t
        \end{pmatrix} \in \RR^{d+t}$
        \\
        $U_t$ & $x^{*\top} V_t^{-1} \Phi_t$
        \\
        \bottomrule
    \end{tabular}
\end{table*}
\begin{table*}[t!]
    \centering
    \caption{Generic notations}
    \label{table:generic notations}
    \begin{tabular}{ c l }
        \toprule
        Sets and functions
        \\
        \hline
        $\NN$ & Set of natural numbers, starting from $1$
        \\
        $[M]$ & Set of natural numbers up to $M$, i.e., $\left\{ 1, 2, \ldots, M\right\}$
        \\
        $\ind$ & Indicator function
        \\
        \hline
        Vector and matrices
        \\
        \hline
        $\| \cdot \|_2$ & $\ell_2$ norm of a vector
        \\
        $ ( \cdot )_i $ & $i$-th element of a vector
        \\
        $\boldsymbol{0}_d$ & Zero vector in $\RR^d$
        \\
        $I_d$ & Identity matrix in $\RR^{d \times d}$
        \\
        \hline
        Probability
        \\
        \hline
        $ \left( \Omega, \Fcal, \PP \right)$ & Probability space
        \\
        $\EE$ & Expectation
        \\
        $\Ncal(\mu, \Sigma)$ & Gaussian distribution with mean $\mu$ and covariance $\Sigma$
        \\
        $\land$ & Logical conjunction (``and'')
        \\
        $\lor$ & Logical disjunction (``or'')
        \\
        \bottomrule
    \end{tabular}
\end{table*}

\section{Proof of Theorem~\ref{thm:general}}
\label{appx:Proof of theorem 3}

\begin{lemma}[Lemmas 10 and 11 in~\citet{abbasi2011improved}]
\label{lma:elliptical potential lemma}
    Let $\lambda \ge 1$, $\left\{ X_t \right\}_{t=1}^T$ be any sequence of $d$-dimensional vectors such that $\| X_t \|_2 \le 1$ for all $t \in [T]$, and $V_t = \lambda I + \sum_{i=1}^t X_i X_i^\top$. 
    Then, $\sum_{t=1}^T \left\| X_t \right\|_{V_{t-1}^{-1}}^2 \le 2 d \log \left( 1 + \frac{T}{d \lambda} \right)$.
\end{lemma}

As mentioned in the discussion of \cref{thm:general}, we utilize Markov's inequality and the random variable of interest is $X_t^\top \theta_t \ind \left\{ \Ecal_{1, t} \cap \Ecal_{2, t} \right \}$.
However, this random variable may not be non-negative, precluding the use of Markov's inequality.
The following lemma shows that adding an appropriate term guarantees its non-negativity.

\begin{lemma}
\label{lma:g is nonnegative}
    Assume the conditions of \cref{thm:general}.
    Let $J(\theta) = \sup_{x \in \Xcal} x^\top \theta$, where $\theta \in \RR^d$.
    Let $\Theta_t = \{ \theta \in \RR^d \mid \| \theta - \theta^* \|_{V_{t-1}} \le \gamma \}$.
    Define $\theta_t^- = \argmin_{\theta \in \Theta_t} J(\theta)$ and $X_t^- = \argmax_{x \in \Xcal} x^\top \theta_t^-$.
    For any $\theta \in \RR^d$ and an event $\Ecal'$, we introduce the following notation:
    \begin{equation*}
        g_t \left( \theta, \Ecal' \right) = \left( J(\theta) - J(\theta_t^-) \right) \ind \{ \Ecal' \}
        \, .
    \end{equation*}
    Then, $g_t(\theta^*, \Ecal' ) \ge 0$ holds for any event $\Ecal' \in \Fcal$, and $g_t(\theta_t, \Ecal'') \ge 0$ holds almost surely for any event such that $\Ecal'' \subset \Ecal_{1, t}$.
\end{lemma}
\begin{proof}
    We first prove $g_t(\theta^*, \Ecal') \ge 0$.
    Since $\theta^* \in \Theta_t$, 
    \begin{equation*}
        J(\theta^*) \ge \inf_{\theta \in \Theta_t} J(\theta) = J(\theta_t^-)
        \, 
    \end{equation*}
    always holds.
    Therefore, for any event $\Ecal'$, $g_t(\theta^*, \Ecal') \ge 0$ holds.
    \\
    We now suppose $\Ecal'' \subset \Ecal_{1, t}$ and prove $g_t(\theta_t, \Ecal'') \ge 0$.
    We consider two cases where ${\Ecal''}$ does and does not hold.
    Under $\Ecal''$, since $\Ecal'' \subset \Ecal_{1, t}$ and by concentration condition~\eqref{eq:Condition 1 concentration}, $\theta_t \in \Theta_t$ holds almost surely.
    Then, $J(\theta_t) \ge \inf_{\theta\in\Theta_t} J(\theta) = J(\theta_t^-)$ holds.
    Under ${\Ecal''}^\Csf$, $g_t(\theta_t, \Ecal'') = 0 \ge 0$ trivially holds.
    Therefore, $g_t(\theta_t, \Ecal'') \ge 0$ holds almost surely.
\end{proof}

\begin{proof}[Proof of \cref{thm:general}]
    We show that with probability at least $1 - \delta$, it holds that
    \begin{equation*}
        R(T) \ind \left\{ \Ecal \right\} \le \gamma \left( 1 + \frac{2}{p} \right) \sqrt{ 2 d T \log \left( 1 + \frac{T}{d\lambda} \right) } + \frac{\gamma}{p} \sqrt{ \frac{2T}{\lambda} \log \frac{1}{\delta}}
        \, .
    \end{equation*}
    We first bound the instantaneous regret under the event $\Ecal$ at time $t$.
    \begin{align*}
        \left( x^{*\top} \theta^* - X_t^\top \theta^* \right) \ind\left\{ \Ecal \right\}
        & = \left( x^{*\top} \theta^* - X_t^\top \theta_t + X_t^\top \theta_t - X_t^\top \theta^*\right) \ind \left\{ \Ecal \right\}
        \\
        & = \underbrace{\left(x^{*\top} \theta^* - X_t^\top \theta_t \right) \ind \left\{ \Ecal \right\}}_{I_1} + \underbrace{\left(X_t^\top \theta_t - X_t^\top \theta^*\right) \ind \left\{ \Ecal \right\}}_{I_2}
    \end{align*}
    $I_2$ is directly bounded under $\Ecal$.
    \begin{align*}
        I_2 & = X_t^\top \left( \theta_t - \theta^*\right) \ind \left\{ \Ecal \right\}
        \\
        & \le \left\| X_t \right\|_{V_{t-1}^{-1}} \left\| \theta_t - \theta^* \right\|_{V_{t-1}}\ind \left\{ \Ecal \right\}
        \\
        & \le \gamma \left\| X_t \right\|_{V_{t-1}^{-1}}
        \, ,
    \end{align*}
    where the first inequality is due to the Cauchy-Schwarz inequality and the second inequality comes from condition~\eqref{eq:Condition 1 concentration}.
    \\
    Now, we bound $I_1$.
    Let $X_t^-$, $\theta_t^-$, and $g_t$ be defined as in \cref{lma:g is nonnegative}.
    By \cref{lma:g is nonnegative}, $g_t (\theta_t, \Ecal) \ge 0$ holds almost surely.
    Then,
    \begin{align*}
        I_1 & = x^{*\top}\theta^* \ind \left\{ \Ecal \right\} - X_t^\top \theta_t \ind \left\{ \Ecal \right\} 
        \\
        & = x^{*\top}\theta^* \ind \left\{ \Ecal \right\} - X_t^{-\top}\theta_t^- \ind \left\{ \Ecal \right\} +  X_t^{-\top}\theta_t^- \ind \left\{ \Ecal \right\} - X_t^\top \theta_t \ind \left\{ \Ecal \right\} 
        \\
        & = g_t\left( \theta^*, \Ecal \right) - g_t\left( \theta_t, \Ecal \right)
        \\
        & \le g_t\left( \theta^*, \Ecal \right)
        \\
        & \le g_t\left( \theta^*, \Ecal_{2, t} \right)
        \, ,
    \end{align*}
    where the last inequality holds since $\Ecal \subset \Ecal_{2, t}$.
    Again by \cref{lma:g is nonnegative}, $g_t\left( \theta^*, \Ecal_{2, t} \right)$ and $g_t(\theta_t, \Ecal_{1, t} \cap \Ecal_{2, t} )$ are non-negative almost surely.
    Note that by the first part of condition~\eqref{eq:Condition 2 Optimism}, $\Ecal_{2, t}$ is $\Fcal_{t-1}$-measurable and hence $g_t\left( \theta^*, \Ecal_{2, t} \right) = ( x^{*\top} \theta^* - X_t^{-\top} \theta_t^- ) \ind \{\Ecal_{2, t} \}$ is also $\Fcal_{t-1}$-measurable.
    Applying Markov's inequality conditioned on $\Fcal_{t-1}$, we obtain that
    \begin{align}
        & g_t(\theta^*, \Ecal_{2, t} )
        \PP \Bigl( g_t(\theta^*, \Ecal_{2, t} ) \le g_t\left( \theta_t, \Ecal_{1, t} \cap \Ecal_{2, t} \right) \mid \Fcal_{t-1} \Bigr)  \le \EE \Bigl[ g_t\left( \theta_t, \Ecal_{1, t} \cap \Ecal_{2, t} \right) \mid \Fcal_{t-1} \Bigr]
        \label{eq:use of markov}
        \, .
    \end{align}
    We lower-bound the probability on the left-hand side utilizing condition~\eqref{eq:Condition 2 Optimism}.
    Suppose the event of interest in condition~\eqref{eq:Condition 2 Optimism}, namely $ ( ( x^{*\top} \theta^* \le X_t^\top \theta_t ) \land \Ecal_{1, t} ) \lor \Ecal_{2, t}^{\Csf}$, holds.
    Under the event, either $\Ecal_{2, t}^{\Csf}$ or $( x^{*\top} \theta^* \le X_t^\top \theta_t ) \land \Ecal_{1, t} \land \Ecal_{2, t}$ holds.
    Under $\Ecal_{2, t}^{\Csf}$, $g_t(\theta^*, \Ecal_{2, t} ) \le g_t\left( \theta_t, \Ecal_{1, t} \cap \Ecal_{2, t} \right) $ becomes $0 \le 0$, which trivially holds.
    Otherwise, we have $( x^{*\top} \theta^* \le X_t^\top \theta_t ) \land \Ecal_{1, t} \land \Ecal_{2, t}$.
    Under this event, we have that
    \begin{align*}
        x^{*\top} \theta^* \le X_t^\top \theta_t
        & \Leftrightarrow
        x^{*\top} \theta^* - X_t^{-\top} \theta_t^-  \le X_t^\top \theta_t  - X_t^{-\top} \theta_t^- 
        \\
        & \Leftrightarrow 
        \left( x^{*\top} \theta^* - X_t^{-\top} \theta_t^-\right) \ind \left\{ \Ecal_{2, t} \right\} \le \left( X_t^\top \theta_t  - X_t^{-\top} \theta_t^- \right) \ind \left\{ \Ecal_{1, t} \cap \Ecal_{2, t} \right\}
        \\
        & \Leftrightarrow
        g_t( \theta^*, \Ecal_{2, t} ) \le g_t( \theta_t, \Ecal_{1, t} \cap \Ecal_{2, t} )
        \, .
    \end{align*}
    Therefore, we have shown that
    \begin{align*}
        \left( \left( x^{*\top} \theta^* \le X_t^\top \theta_t \right) \land \Ecal_{1, t} \right) \lor \Ecal_{2, t}^{\Csf}
        \Rightarrow
        g_t( \theta^*, \Ecal_{2, t} ) \le g_t( \theta_t, \Ecal_{1, t} \cap \Ecal_{2, t} )
        \, ,
    \end{align*}
    which implies
    \begin{align*}
        \PP \left( g_t( \theta^*, \Ecal_{2, t} ) \le g_t\left( \theta_t, \Ecal_{1, t} \cap \Ecal_{2, t} \right) \mid \Fcal_{t-1} \right)
        & \ge \PP \left( \left( x^{*\top} \theta^* \le X_t^\top \theta_t \land \Ecal_{1, t} \right) \lor \Ecal_{2, t}^{\Csf} \mid \Fcal_{t-1} \right)
         \\
         & \ge p
         \,
    \end{align*}
    by condition~\eqref{eq:Condition 2 Optimism}.
    Therefore, we obtain that ${g_t( \theta^*, \Ecal_{2, t}) \le \frac{1}{p} \EE [ g_t( \theta_t, \Ecal_{1, t} \cap \Ecal_{2, t}) \mid \Fcal_{t-1} ]}$ from inequality~\eqref{eq:use of markov}.
    Lastly, we bound $g_t( \theta_t, \Ecal_{1, t} \cap \Ecal_{2, t} )$.
    \begin{align*}
        g_t \left( \theta_t, \Ecal_{1, t} \cap \Ecal_{2, t} \right)
        &= \left( X_t^\top \theta_t - X_t^{-\top} \theta_t^- \right) \ind \left\{ \Ecal_{1, t} \cap \Ecal_{2, t} \right\}
        \\
        & \le \left( X_t^\top \theta_t - X_t^{\top} \theta_t^- \right) \ind \left\{ \Ecal_{1, t} \cap \Ecal_{2, t} \right\}
        \\
        & \le \left\| X_t \right\|_{V_{t-1}^{-1}} \left\| \theta_t - \theta_t^{-} \right\|_{V_{t-1}} \ind \left\{ \Ecal_{1, t} \right\}
        \\
        & \le \left\| X_t \right\|_{V_{t-1}^{-1}} \left( \left\| \theta_t - \theta^* \right\|_{V_{t-1}} + \left\| \theta_t^- - \theta^* \right\|_{V_{t-1}} \right) \ind \left\{ \Ecal_{1, t} \right\}
        \\
        & \le 2 \gamma \left\| X_t \right\|_{V_{t-1}^{-1}}
        \, ,
    \end{align*}
    where the first inequality uses that $X_t^- = \sup_{x \in \Xcal} x^\top \theta_t^-$, the second inequality is due to the Cauchy-Schwarz inequality, the third inequality holds by the triangle inequality, and the last inequality comes from condition~\eqref{eq:Condition 1 concentration} and that $\theta_t^- \in \Theta_t$ as defined in \cref{lma:g is nonnegative}.
    Combining all, the instantaneous regret at time $t$ under the event $\Ecal$ is bounded as follows:
    \begin{align*}
        \left( x^{*\top} \theta^* - X_t^\top \theta^* \right) \ind\left\{ \Ecal \right\}
        & \le \gamma \left\| X_t \right\|_{V_{t-1}^{-1}} + \frac{2 \gamma }{p} \EE \left[ \left\| X_t \right\|_{V_{t-1}^{-1}} \mid \Fcal_{t-1} \right]
        \\
        & = \left( \gamma + \frac{2 \gamma}{p} \right) \left\| X_t \right\|_{V_{t-1}^{-1}} + \frac{2 \gamma}{p} \left( \EE \left[ \left\| X_t \right\|_{V_{t-1}^{-1}} \mid \Fcal_{t-1} \right] - \left\| X_t \right\|_{V_{t-1}^{-1}}\right) 
        \, .
    \end{align*}
    Then, the cumulative regret under $\Ecal$ is bounded as
    \begin{align*}
        R(T) \ind \left\{ \Ecal \right\}
        & \le 
        \gamma \left( 1 + \frac{2}{p} \right) \sum_{t=1}^T \left\| X_t \right\|_{V_{t-1}^{-1}} + \frac{2 \gamma}{p} \sum_{t=1}^T \left( \EE \left[ \left\| X_t \right\|_{V_{t-1}^{-1}} \mid \Fcal_{t-1} \right] - \left\| X_t \right\|_{V_{t-1}^{-1}} \right)
        \, .
    \end{align*}
    To bound the first sum, we apply the Cauchy-Schwarz inequality and then \cref{lma:elliptical potential lemma}.
    \begin{align*}
        \sum_{t=1}^T \left\| X_t \right\|_{V_{t-1}^{-1}}
        & \le \sqrt{ T \sum_{t=1}^T \left\| X_t \right\|_{V_{t-1}^{-1}}^2 }
        \\
        & \le \sqrt{ 2 d T \log \left( 1 + \frac{T}{d \lambda} \right)}
        \, .
    \end{align*}
    The second sum is bounded by the Azuma-Hoeffding inequality. 
    Note that $0 \le \left\| X_t \right\|_{V_{t-1}^{-1}} \le \sqrt{\lambda_{\max}(V_{t-1}^{-1})} \left\| X_t \right\|_2 \le \frac{1}{\sqrt{\lambda}}$, where $\lambda_{\max}( \cdot )$ denotes the maximum eigenvalue.
    By \cref{lma:Azuma-Hoeffding inequality}, with probability at least $1 - \delta$, it holds that
    \begin{equation*}
        \sum_{t=1}^T \left( \EE \left[ \left\| X_t \right\|_{V_{t-1}^{-1}} \mid \Fcal_{t-1} \right] - \left\| X_t \right\|_{V_{t-1}^{-1}} \right)
        \le \sqrt{ \frac{T}{2 \lambda} \log \frac{1}{\delta}} \, .
    \end{equation*}
    Therefore, with probability at least $1 - \delta$, the cumulative regret under $\Ecal$ is bounded as follows:
    \begin{equation*}
        R(T) \ind \left\{ \Ecal \right\} \le \gamma \left( 1 + \frac{2}{p} \right) \sqrt{ 2 d T  \log \left( 1 + \frac{T}{d\lambda} \right) } + \frac{\gamma}{p} \sqrt{ \frac{2T}{\lambda} \log \frac{1}{\delta}}
        \, .
    \end{equation*}
\end{proof}

\section{Proof of Lemma~\ref{lma:optimism lemma}}
\label{appx:Proof of optimism lemma}
\begin{proof}[Proof of \cref{lma:optimism lemma}]
    The proof is simple algebra utilizing a useful matrix $\Phi_t$.
    Define $\Phi_t$ to be the matrix that stacks $X_i$ in addition to an identity matrix as follows:
    \begin{align*}
        \Phi_t &:= \begin{pmatrix}
            \sqrt{\lambda} I_d &  X_1 & \cdots & X_t
        \end{pmatrix} \in \RR^{d \times (d + t)}
        \, .
    \end{align*}
    We can express a relevant matrix and vectors using $\Phi_t$, which are $V_t = \Phi_t \Phi_t^\top$, $\tilde{\theta}_{t}^j = V_{t}^{-1} \Phi_{t} \Zb_{t+1}^j$, and $U_t^\top = x^{*\top} V_{t}^{-1} \Phi_t$.
    Defining $\tilde{\theta}_{t-1} = \tilde{\theta}_{t-1}^{j_t}$ to be the perturbation in the selected estimator at time $t$, we obtain that 
    \begin{align*}
        x^{*\top} \tilde{\theta}_{t-1} & = x^{*\top} V_{t-1}^{-1} \Phi_{t-1} {\Zb}_t
        \\
        & = U_{t-1}^\top {\Zb}_t
        \, .
    \end{align*}
    In addition, it holds that 
    \begin{align*}
        \left\| U_{t-1} \right\|_2
        & = \sqrt{x^{*\top} V_{t-1}^{-1}  {\Phi}_{t-1}{\Phi}_{t-1}^\top V_{t-1}^{-1} x^* }
        \\
        & = \sqrt{x^{*\top} V_{t-1}^{-1} x^* }
        \\
        & = \left\| x^* \right\|_{V_{t-1}^{-1}}
        \, .
    \end{align*}
    By $X_t^\top \theta_t = \sup_{x \in \Xcal} x^\top \theta_t$, it holds that $x^{*\top} \theta_t \le X_t^\top \theta_t$.
    Then, we have that
    \begin{align*}
        X_t^\top \theta_t - x^{*\top} \theta^*
        & \ge x^{*\top} \theta_t - x^{*\top} \theta^*
        \\
        & = x^{*\top} \left( \theta_t - \theta^* \right)
        \\
        & = x^{*\top} \left( \tilde{\theta}_{t-1} + \hat{\theta}_{t-1} - \theta^* \right)
        \\
        & = U_{t-1} \Zb_t  + x^{*\top} \left( \hat{\theta}_{t-1} - \theta^* \right)
        \, .
    \end{align*}
    By the condition of the lemma, there exists a positive constant $c$ such that $\| \hat{\theta}_{t-1} - \theta^* \|_{V_{t-1}} \le c$ and $U_{t-1}^\top \Zb_t - c \| U_{t-1} \|_2 \ge 0$.
    By the Cauchy-Schwarz inequality, it holds that
    \begin{align*}
        x^{*\top} \left( \hat{\theta}_{t-1} - \theta^* \right) 
        & \ge - \left\| x^* \right\|_{V_{t-1}^{-1}} \bigl\| \hat{\theta}_{t-1} - \theta^* \bigr\|_{V_{t-1}} 
        \\
        & \ge - c \left\| U_{t-1} \right\|_2
        \, .
    \end{align*}
    Therefore, 
    \begin{align*}
        X_t^\top \theta_t - x^{*\top}\theta^*
        & \ge  U_{t-1}^\top {\Zb}_t - c \left\| U_{t-1} \right\|_2
        \\
        & \ge 0
        \, ,
    \end{align*}
    where we proved that $X_t^\top \theta_t \ge x^{*\top} \theta^*$.
\end{proof}

\section{Proof of Lemma~\ref{lma:rpe result}}
\label{appx:Proof of rpe lemma}
\begin{proof}[Proof of \cref{lma:rpe result}]
    Under $\Ecal_{1, t} = \tilde{\Ecal}_{1, t} \cap \hat{\Ecal}_{t-1}$, the concentration condition, specifically condition~\eqref{eq:Condition 1 concentration} in \cref{thm:general}, holds by the triangle inequality.
    \begin{align*}
        \left\| \theta_t - \theta^* \right\|_{V_{t-1}} = \bigl\| \tilde{\theta}_{t-1} + \hat{\theta}_{t-1} - \theta^* \bigr\|_{V_{t-1}} \le \bigl\| \tilde{\theta}_{t-1} \bigr\|_{V_{t-1}} + \bigl\| \hat{\theta}_{t-1} - \theta^* \bigr\|_{V_{t-1}} \le \tilde{\gamma} + \beta_{t-1} \le \gamma
        \, .
    \end{align*}
    \\
    To show the optimism condition, condition~\eqref{eq:Condition 2 Optimism} in \cref{thm:general}, we first show that $\Ecal_{2, t} \in \Fcal_{t-1}$.
    ${\hat{\Ecal}_{t-1} \in \Fcal_{t-1}}$ holds since it regards $\left\{ X_i, \eta_i \right\}_{i=1}^{t-1}$ only.
    Since $\PP ( (U_{t-1}^\top {\Zb}_t \ge \beta_{t-1} \left\| U_{t-1}\right\|_2 ) \land \tilde{\Ecal}_{1, t} \mid \Fcal_{t-1} )$ is a $\Fcal_{t-1}$-measurable random variable and $\tilde{\Ecal}_{2, t}$ is an event that the specified random variable is greater than or equal to $p$, $\tilde{\Ecal}_{2, t}$ is in $\Fcal_{t-1}$.
    Therefore, we obtain that $\Ecal_{2, t} = \hat{\Ecal}_{t-1} \cap \tilde{\Ecal}_{2, t} \in \Fcal_{t-1}$.
    To prove the remaining part of condition~\eqref{eq:Condition 2 Optimism}, we demonstrate the following logical implication relationships:
    \begin{align*}
        \left( \left( U_{t-1}^\top {\Zb}_t \ge \beta_{t-1} \left\| U_{t-1} \right\|_2 \right) \land \tilde{\Ecal}_{1, t} \right) \lor \Ecal_{2, t}^\Csf
        & \Rightarrow
        \left( \left( U_{t-1}^\top {\Zb}_t \ge \beta_{t-1} \left\| U_{t-1} \right\|_2  \right) \land \tilde{\Ecal}_{1, t} \land \Ecal_{2, t} \right) \lor \Ecal_{2, t}^\Csf
        \\
        & \Rightarrow
        \left( \left( U_{t-1}^\top {\Zb}_t \ge \beta_{t-1} \left\| U_{t-1} \right\|_2  \right) \land \tilde{\Ecal}_{1, t} \land \hat{\Ecal}_{t-1} \right) \lor \Ecal_{2, t}^\Csf
        \\
        & \Rightarrow
        \left( \left( x^{*\top} \theta^* \le X_t^\top \theta_t \right) \land \tilde{\Ecal}_{1, t} \land \hat{\Ecal}_{t-1} \right) \lor \Ecal_{2, t}^\Csf
        \\
        & \Rightarrow
        \left( \left( x^{*\top} \theta^* \le X_t^\top \theta_t \right) \land \Ecal_{1, t} \right) \lor \Ecal_{2, t}^\Csf
        \, ,
    \end{align*}
    where the first implication follows from $A \lor B^\Csf \Leftrightarrow \left( A \land B \right) \lor B^\Csf$, the second implication holds since $\Ecal_{2, t} \subset \hat{\Ecal}_{t-1}$, the third implication holds by \cref{lma:optimism lemma}, which states that $( U_{t-1}^\top {\Zb}_t \ge \beta_{t-1} \left\| U_{t-1} \right\|_2 \land \hat{\Ecal}_{t-1}) \Rightarrow ( x^{*\top} \theta^* \le X_t^\top \theta_t )$, and the last by $\Ecal_{1, t} = \tilde{\Ecal}_{1, t} \cap \hat{\Ecal}_{t-1}$.
    This implication relationship shows that
    \begin{align*}
        \PP \left( \left( \left( x^{*\top} \theta^* \le X_t^\top \theta_t \right) \land \Ecal_{1, t} \right) \lor \Ecal_{2, t}^\Csf \mid \Fcal_{t-1} \right) & \ge 
        \PP \left(\left( \left( U_{t-1}^\top {\Zb}_t \ge \beta_{t-1} \left\| U_{t-1} \right\|_2 \right) \land \tilde{\Ecal}_{1, t} \right) \lor \Ecal_{2, t}^\Csf \mid \Fcal_{t-1} \right)
        \, .
    \end{align*}
    We bound the right-hand side using the definition of $\Ecal_{2, t}$.
    \begin{align*}
        & \PP \left(\left( \left( U_{t-1}^\top {\Zb}_t \ge \beta_{t-1} \left\| U_{t-1} \right\|_2 \right) \land \tilde{\Ecal}_{1, t} \right) \lor \Ecal_{2, t}^\Csf \mid \Fcal_{t-1} \right)
        \\
        & = \EE \left[ \ind \left\{ \left( \left( U_{t-1}^\top {\Zb}_t \ge \beta_{t-1} \left\| U_{t-1} \right\|_2 \right) \land \tilde{\Ecal}_{1, t} \right) \lor \Ecal_{2, t}^\Csf \right\} \mid \Fcal_{t-1} \right]
        \\
        & = \EE \left[ \ind \left\{\left( U_{t-1}^\top {\Zb}_t \ge \beta_{t-1} \left\| U_{t-1} \right\|_2 \right) \land \tilde{\Ecal}_{1, t} \right\} \ind \left\{ \Ecal_{2, t} \right\} + \ind \left\{ \Ecal_{2, t}^{\Csf} \right\} \mid \Fcal_{t-1} \right]
        \\
        & = \EE \left[ \ind \left\{  \left( U_{t-1}^\top {\Zb}_t \ge \beta_{t-1} \left\| U_{t-1} \right\|_2 \right) \land \tilde{\Ecal}_{1, t}  \right\} \mid \Fcal_{t-1} \right] \ind \left\{ \Ecal_{2, t} \right\} + \ind \left\{ \Ecal_{2, t}^{\Csf} \right\} 
        \\
        & \ge p \ind \left\{ \Ecal_{2, t} \right\} + \ind \left\{ \Ecal_{2, t}^\Csf \right\}
        \\
        & \ge p
        \, ,
    \end{align*}
    where the third equality uses that $\Ecal_{2, t} \in \Fcal_{t-1}$ and the first inequality holds since under $\Ecal_{2, t} \subset \tilde{\Ecal}_{2, t}$, $\EE \left[ \ind \left\{ \left(  U_{t-1}^\top {\Zb}_t \ge \beta_{t-1} \left\| U_{t-1} \right\|_2 \right) \land \tilde{\Ecal}_{1, t} \right\} \mid \Fcal_{t-1} \right] \ge p$ holds by the definition of $\tilde{\Ecal}_{2, t}$.
    \\
    Therefore, $\Ecal_{1, t}$ and $\Ecal_{2, t}$ satisfy conditions~\eqref{eq:Condition 1 concentration}~and~\eqref{eq:Condition 2 Optimism}.
    By \cref{thm:general}, the regret bound stated in inequality~\eqref{eq:regret bound} holds with probability at least $1 - \delta - \PP (\Ecal^\Csf)$, where the union bound is taken.
    Note that $\Ecal = \cap_{t=1}^T \left(\Ecal_{1, t} \cap \Ecal_{2, t} \right) = \cap_{t=1}^T \left( \tilde{\Ecal}_{1, t} \cap \tilde{\Ecal}_{2, t} \cap \hat{\Ecal}_{t-1} \right) \supset \hat{\Ecal} \cap \tilde{\Ecal}$.
    The failure probability is bounded as
    \begin{align*}
        \delta + \PP \left( \Ecal^\Csf \right) 
        & \le \delta + \PP\left(\hat{\Ecal}^\Csf\right) + \PP \left( \tilde{\Ecal}^\Csf \right)
        \\
        & \le 2\delta +  \PP \left( \tilde{\Ecal}^\Csf \right)
        \, ,
    \end{align*}
    where the first inequality takes the union bound over $\Ecal^\Csf \subset \hat{\Ecal}^\Csf \cup \tilde{\Ecal}^\Csf$ and the second inequality is due to \cref{lma:theta hat concetration}.
\end{proof}

\section{Proof of Lemma~\ref{lma:gaussian concentration lemma}}
\label{appx:Proof of gaussian concentration}
\cref{lma:gaussian concentration lemma} is a special case of \cref{lma:concentration lemma}, which generalizes Gaussian distribution to any subGaussian distribution.
We first provide a general chi-squared concentration result, which is required to bound the perturbation induced by $W$.
A generalized version of \cref{lma:Chi squared tail inequality} is presented in \cref{lma:bound 2norm}.
\begin{lemma}
\label{lma:Chi squared tail inequality}
    If $Z \sim \Ncal(0, I_d)$ is a $d$-dimensional multivariate Gaussian vector, then for any $\delta \in (0, 1]$,
    \begin{equation*}
        \PP \left( \left\| Z \right\|_2 \ge \sqrt{d} + \sqrt{2\log \frac{1}{\delta}} \right) \le \delta
        \, .
    \end{equation*}
\end{lemma}
\begin{proof}
    By \cref{lma:chi-squared} with $x = \log \frac{1}{\delta}$, it holds that
    \begin{equation*}
        \PP \left( \left\| Z \right\|_2^2 - d \ge 2 \sqrt{d \log \frac{1}{\delta}} + 2 \log \frac{1}{\delta} \right) \le \delta
        \, .
    \end{equation*}
    Since $d + 2 \sqrt{d \log \frac{1}{\delta}} + 2 \log \frac{1}{\delta} \le \left(\sqrt{d} + \sqrt{2 \log \frac{1}{\delta}} \right)^2$, it holds that
    \begin{align*}
        \PP \left( \left\| Z \right\|_2 \ge \sqrt{d} + \sqrt{2\log \frac{1}{\delta}}\right)
        & \le \PP \left( \left\| Z \right\|_2^2 \ge d + 2 \sqrt{d \log \frac{1}{\delta}} + 2 \log \frac{1}{\delta} \right)
        \\
        & \le \delta
        \, .
    \end{align*}
\end{proof}

\begin{proof}[Proof of \cref{lma:gaussian concentration lemma}]
    By \cref{lma:Chi squared tail inequality}, with $\delta / 2T$ instead of $\delta$, yields
    \begin{align*}
        \PP \left( \left\| W^j \right\|_2 \ge \sqrt{\lambda} \beta_T \left( \sqrt{d} + \sqrt{2 \log \frac{2T}{\delta}} \right) \right) \le \frac{\delta}{2T}
        \, ,
    \end{align*}
    since $W^j \sim \Ncal(\boldsymbol{0}_d, \lambda \beta_T^2 I_d)$.
    Applying \cref{lma:concentration lemma} with $\delta / 2T$ instead of $\delta$ yields that
    \begin{equation*}
        \bigl\| \tilde{\theta}_{t-1}^j \bigr\|_{V_{t-1}} \le \beta_T \sqrt{ d \log \left( 1 + \frac{T}{d \lambda} \right) + 2 \log \frac{2T}{\delta} } + \beta_T \left( \sqrt{d} + \sqrt{ 2 \log \frac{2T }{\delta} } \right)
        \,
    \end{equation*}
    holds with probability at least $1 - \delta / T$.
    Note that the right-hand side is equal to the definition of $\tilde{\gamma}_T$, defined in Eq.~\eqref{eq:definition of gamma_T}.
\end{proof}

\section{Rigorous Justification of Proposition~\ref{prop:optimism}}
\label{appx:rigorous explanation}
In this section, we rigorously justify \cref{prop:optimism} that is stated in the proof of \cref{thm:enssamp}.
To do so, we present a different viewpoint on the perturbation sequences.

Denote the arms as $\Xcal = \{ x^1, x^2, \ldots, x^K\}$.
For the sake of the analysis, assume that $\delta / T \le p_N / 2 \approx 0.08$, which holds whenever $T \ge 14$ or $\delta < 0.07$.
\\
We reconstruct the perturbation sampled by \cref{alg:enssamp}.
Assume that in addition to $\{W^j\}_{j=1}^m \stackrel{\iid}{\sim} \Pcal_I$, \cref{alg:enssamp} samples $mKT$ samples of $\{ \{Z_{k, t}^j \}_{(k, t)} \}_{j=1}^m \stackrel{\iid}{\sim} \Pcal_R$ at the beginning, where the subscript $(k, t)$ enumerates from $(1, 1)$ to $(K, T)$.
Define $N_{k, t} = \sum_{i=1}^t \ind \left\{ X_i = x^k \right\}$ to be the number of times arm $k$ has been chosen up to time $t$.
If the $a_t$-th arm, $x^{a_t}$, is selected at time $t$, then we assign $Z_t^j = Z_{a_t, N_{a_t, t}}^j$.
Since $Z_{a_t, N_{a_t, t}}^j$ is still an i.i.d. sample of $\Pcal_R$ conditioned on history, specifically on $\sigma ( \Fcal_t^X \cup \Fcal_t^\eta \cup \sigma ( \{ \{ Z_i^j \}_{i=1}^{t-1} \}_{j=1}^m ) )$, we attain an equivalent algorithm with \cref{alg:enssamp}.
We note that these modifications need not be taken in the execution of the algorithm, and their purpose is purely for the analysis.
Define $\Fcal_0^A = \sigma ( \{ W^j, \{Z_{k, t}^j \}_{(k, t)} \}_{j=1}^m )$, which reflects the fact that they are sampled in advance.
For $t \in [T]$, define $\Fcal_t^A = \sigma \left( \Fcal_{t-1} \cup \sigma( j_t ) \right)$, which indicates that the only additional randomness of the algorithm when choosing $\theta_t$ is the sampling of $j_t \sim \Jcal_t$.
Define the extended perturbation vector as follows:
\begin{equation*}
    \Zb_{KT}^j = \begin{pmatrix}
        \frac{1}{\sqrt{\lambda}} {W^j}^\top & Z_{1, 1}^j & \ldots & Z_{1, T}^j & Z_{2, 1}^j & \ldots & Z_{K, T}^j
    \end{pmatrix}^\top \in \RR^{d + KT}
\end{equation*}
Removing some components and reordering ${\Zb}_{KT}^j$ yields ${\Zb}_t^j$, where the removal and reordering depend on the sequence of chosen arms, namely $a_1, \ldots, a_{t-1}$.
Note that $\Zb_{KT}^j \sim \Ncal( \boldsymbol{0}_{d + KT}, \beta_T^2 I_{d + KT})$.
We also define the corresponding extensions of $\Phi_t$ and $U_t$.
Define a matrix that has $n$ columns, first $a$ of which are copies of $v \in \RR^d$ and the rest are $\boldsymbol{0}_d$ as follows.
\begin{equation*}
    \mathrm{rep}(v, a, n) := \begin{pmatrix}
        v & \ldots & v
        & \boldsymbol{0}_d & \ldots & \boldsymbol{0}_d
    \end{pmatrix}
    \in \RR^{d \times n}
    \, .
\end{equation*}
Define the extended version of $\Phi_t$ as follows:
\begin{equation*}
    \Phib_t = \begin{pmatrix}
        \sqrt{\lambda} I_d
        & \mathrm{rep}(x^1, N_{1, t}, T) & \ldots & \mathrm{rep}(x^K, N_{K, t}, T)
    \end{pmatrix} 
    \in \RR^{d \times (d +  KT)}
    \, .
\end{equation*}
$\Phib_t$ extends $\Phi_t$ by permuting the columns so that the feature vectors from the same arm appear in consecutive columns, then inserting multiple $\boldsymbol{0}_d$ appropriately.
Then, we have $V_t = \Phib_t \Phib_t^\top$ and $\tilde{\theta}_t = V_t^{-1} \Phib_t \Zb^j_{KT}$, since $\Zb_{KT}^j$ is permuted and extended from $\Zb_t^j$ in a similar manner.
We define the extended version of $U_t$ as $\Ub_{t} = ( x^{*\top} V_{t}^{-1} \Phib_t )^\top\in \RR^{d + KT}$.
$\Ub_t$ is also a permutation of $U_t$ with additional zeros inserted.
It holds that $U_t^\top {\Zb}_t^j = \Ub_t^\top \Zb_{KT}^j$ and $\left\| U_t \right\|_2 = \left\| \Ub_t \right\|_2$.
\\
Let $\XX_t$ be the set of all possible $\Phib_t$.
Since $\Phib_t$ is fully determined by $N_{1, t}, \ldots N_{K, t}$ and each $N_{k, t}$ takes value between 0 and $T - 1$ inclusively when $0 \le t \le T-1$, we obtain that $\left| \cup_{t=0}^{T-1} \XX_t \right| \le T^K$.
For any $t \in [T]$, take any $\Phib_{t-1} \in \XX_{t-1}$.
Note that
$\Ub_{t-1} = x^{*\top} (\Phib_{t-1} \Phib_{t-1}^\top)^{-1} \Phib_{t-1}$ is fully determined by $\Phib_{t-1}$, and $\tilde{\theta}_{t-1}^j = (\Phib_{t-1} \Phib_{t-1}^\top)^{-1} \Phib_{t-1} \Zb_{KT}^j$ is determined by $\Phib_{t-1}$ and $\Zb_{KT}^j$.
Assuming that $\Phib_{t-1}$ is fixed, $\Ub_{t-1}$ is also fixed, therefore we can apply \cref{fact:gaussian} and obtain that $\PP \big( \Ub_{t-1}^\top \Zb_{KT}^j \ge \beta_T \| \Ub_{t-1} \|_2 \big) \ge p_N$, where the only source of randomness comes from $\Zb_{KT}^j$.
Applying \cref{lma:gaussian concentration lemma}, we obtain that $\PP ( \| \tilde{\theta}_{t-1}^j \|_{V_{t-1}} \le \tilde{\gamma}_T ) \ge 1 - \delta/ T$.
Let $I^j ( \Phib_{t-1} ) = \ind \{ {( \Ub_{t-1}^\top \Zb_{KT}^j \ge \beta_T \| \Ub_{t-1} \|_2 ) }\land ( \| \tilde{\theta}_{t-1}^j \|_{V_{t-1}} \le \tilde{\gamma}_T ) \}$.
Then, $\PP ( I^j (\Phib_{t-1}) = 1 ) \ge p_N - \delta / T \ge p_N / 2$.
Since the only randomness on choosing the arm conditioned on $\Fcal_{t-1}$ comes from sampling $j_t$, it holds that
\begin{align*}
    \PP \left( \left( \Ub_{t-1}^\top \Zb_{KT}^{j_t} \ge \beta_T \left\| \Ub_{t-1} \right\|_2 \right) \land \left( \bigl\| \tilde{\theta}_{t-1}^{j_t} \bigr\|_{V_{t-1}} \le \tilde{\gamma}_T \right) \mid \Fcal_{t-1} \right) = \frac{1}{m} \sum_{j=1}^m I^j\left( \Phib_{t-1} \right)
    \, .
\end{align*}
We apply the Azuma-Hoeffding inequality to show that $\frac{1}{m} \sum_{j=1}^m I^j\left( \Phib_{t-1} \right)$ is bounded below with high probability.
Since $\{ I^j(\Phib_{t-1}) \}_{j=1}^m$ are i.i.d. Bernoulli random variables with the associated probability greater than $p_N / 2$, it holds that
\begin{align*}
    & \quad \PP \left( \frac{1}{m} \sum_{j=1}^m I^j(\Phib_{t-1} ) \le \frac{p_N}{4} \right)
    \\
    & = \PP \left( \frac{1}{m} \sum_{j=1}^m I^j(\Phib_{t-1}) - \frac{1}{m} \sum_{j=1}^m \EE[ I^j(\Phib_{t-1}) ] \le \frac{p_N}{4} - \frac{1}{m} \sum_{j=1}^m \EE [ I^j(\Phib_{t-1}) ] \right)
    \\
    & \le \PP \left( \frac{1}{m} \sum_{j=1}^m I^j(\Phib_{t-1}) - \frac{1}{m} \sum_{j=1}^m \EE [ I^j(\Phib_{t-1}) ] \le \frac{p_N}{4} - \frac{p_N}{2} \right) 
    \\
    & = \PP \left( \frac{1}{m} \sum_{j=1}^m I^j(\Phib_{t-1}) - \frac{1}{m} \sum_{j=1}^m \EE [I^j(\Phib_{t-1}) ] \le - \frac{p_N}{4}\right) 
    \\
    & \le \exp \left( - \frac{p_N^2 m}{8} \right)
    \, ,
\end{align*}
where \cref{lma:Azuma-Hoeffding inequality} is applied at the end.
By taking the union bound over $\cup_{t=0}^{T-1} \XX_t$, we obtain that
\begin{align*}
    \PP \left( \exists \Phib \in \cup_{t=0}^{T-1} \XX_{t}, \frac{1}{m} \sum_{j=1}^m I^j(\Phib )  \le \frac{p_N}{4} \right) \le T^K \exp \left( - \frac{p_N^2 m}{8} \right)
    \, .
\end{align*}
The event $\Ecal_2^*$ is defined as the complement of the event above. The proof is complete.
\begin{align*}
    \Ecal_2^* := \left\{ \omega \in \Omega : \forall \Phib \in \cup_{t=0}^{T-1} \XX_t, \frac{1}{m} \sum_{j=1}^m I^j(\Phib) > \frac{p_N}{4} \right\}
    \, .
\end{align*}

\section{Proof of Corollary~\ref{cor:phe}}
\label{sec:proof of phe}

\begin{proof}[Proof of \cref{cor:phe}]
    Let $\tilde{\Ecal}_{1, t}$ and $\tilde{\Ecal}_{2, t}$ be defined as in \cref{lma:rpe result} with $\tilde{\gamma} = \tilde{\gamma}_T$ and $p = p_N / 2$.
    We redefine a couple of notations to adapt \cref{alg:phe}.
    Let 
    \begin{align*}
        \Zb_t := \begin{pmatrix}
        \frac{1}{\sqrt{\lambda}} W_t^\top & Z_{t, 1} & \ldots & Z_{t, t-1} 
    \end{pmatrix}^\top \in \RR^{d + t - 1}
    \end{align*}
    to be the perturbation vector at time $t$, and 
    \begin{align*}
        \tilde{\theta}_{t-1} := V_{t-1}^{-1} \left( W_t +  \sum_{i=1}^{t-1} X_i Z_{t, i} \right)
    \end{align*}
    be the perturbation in the estimator $\theta_t$.
    Regarding $\tilde{\theta}_{t-1}$ as one of $\tilde{\theta}_{t-1}^j$ in the proof of \cref{thm:enssamp}, we obtain that $\PP ( \tilde{\Ecal}_{1, t} ) \ge 1 - \delta / T $ and $\PP ( (U_{t-1}^\top \Zb_t \ge \beta_{t-1} \| U_{t-1} \|_2) \land \tilde{\Ecal}_{1, t} ) \ge p_N / 2$ hold, analogously to inequalities~\eqref{eq:Ecal_1 bound} and \eqref{eq:optimism condition bound} respectively.
    Moreover, in contrast to the proof of \cref{thm:enssamp}, the perturbation vector $\Zb_t$ is now independent of $\Fcal_{t-1}$.
    Noting that $U_{t-1}$ is $\Fcal_{t-1}$-measurable, it always holds that 
    \begin{align*}
        \PP \left( (U_{t-1}^\top \Zb_t \ge \beta_{t-1} \| U_{t-1} \|_2) \land \tilde{\Ecal}_{1, t} \mid \Fcal_{t-1} \right) \ge  \frac{p_N}{2}
        \, .
    \end{align*}
    This proves that $\tilde{\Ecal}_{2, t}$ is in fact the whole event.
    Taking the union bound, we obtain that $\PP ( \tilde{\Ecal}^\Csf ) \le \sum_{t=1}^T \PP ( \tilde{\Ecal}_{1, t}^\Csf ) \le \delta$.
    By \cref{lma:rpe result}, with probability at least $1 - 3 \delta$, the cumulative regret is bounded by
    \begin{align*}
        R(T) \le \gamma_T \left( 1 + \frac{4}{p_N} \right) \sqrt{ 2 d T \log \left( 1 + \frac{T}{d\lambda} \right) } + \frac{2 \gamma_T}{p_N} \sqrt{ \frac{2T}{\lambda} \log \frac{1}{\delta}}
         = \Ocal \left((d \log T) ^{\frac{3}{2}} \sqrt{T}\right) \, .
    \end{align*}    
\end{proof}

\section{Generalizability of Perturbation Distributions}
\label{appx:beyond gaussian}

In this section, we demonstrate that any distribution that is symmetric, subGaussian, and has lower-bounded variance satisfies the results of \cref{lma:gaussian concentration lemma} and \cref{fact:gaussian}, possibly up to a constant factor.
As mentioned in \cref{rmk:non gaussian}, it implies that our results are valid when the Gaussian distribution is replaced with any symmetric non-degenerate subGaussian distribution.
The following lemma is a standard concentration result for vector martingales with subGaussian noises.

\begin{lemma}[Theorem~1 in~\citet{abbasi2011improved}]
\label{lma:self-normalized bound}
    Let $\left\{ \Fcal_t \right\}_{t=0}^\infty$ be a filtration.
    Let $\{ \xi_t \}_{t=1}^\infty$ be a sequence of real-valued random variables such that $\xi_t$ is $\Fcal_t$-measurable and is $\Fcal_{t-1}$-conditionally $\sigma$-subGaussian for some $\sigma \ge 0$.
    Let $\{ X_t \}_{t=1}^\infty$ be a sequence of $\RR^d$-valued random vectors such that $X_t$ is $\Fcal_{t-1}$-measurable and $\left\| X_t \right\|_2 \le 1$ almost surely for all $t \ge 1$.
    Fix $\lambda \ge 1$.
    Let $V_t = \lambda I + \sum_{i=1}^t X_t X_t^\top$.
    Then, for any $\delta \in (0, 1]$, with probability at least $1 - \delta$, the following inequality holds for all $t \ge 0$:
    \begin{equation*}
        \left\| \sum_{i=1}^t \xi_i X_i \right\|_{V_{t}^{-1}} \le \sigma \sqrt{d \log \left( 1 + \frac{t}{d \lambda} \right) + 2 \log \frac{1}{\delta}}
        \, .
    \end{equation*}
\end{lemma}

Next lemma is a simple application of \cref{lma:self-normalized bound}, which proves the concentration result of $\tilde{\theta}_t$ under the subGaussianity of $\Pcal_R$.

\begin{lemma}[Sufficient condition for concentration]
\label{lma:concentration lemma}
    Fix any $\delta \in (0, 1]$ and $t \in [T]$.
    Assume that $\PP \left( \left\| W \right\|_2 > L_{t, 0}^{\delta} \right) \le \delta$, and $\{Z_{i}\}_{i=1}^{t-1}$ are mutually independent of each other and are $\Fcal_i^X$-conditionally $\sigma_R^2$-subGaussian respectively for each $i \in [t-1]$.
    Then, with probability at least $1 - 2 \delta$, it holds that
    \begin{equation*}
        \bigl\| \tilde{\theta}_{t-1} \bigr\|_{V_{t-1}} \le \sigma_R \sqrt{ d \log \left( 1 + \frac{ T }{d \lambda} \right) + 2 \log \frac{1}{\delta} } + \frac{L_{t, 0}^\delta}{\sqrt{\lambda}}
        \, .
    \end{equation*}
\end{lemma}
\begin{proof}
    Recall that $\tilde{\theta}_{t-1} = V_{t-1}^{-1} ( W + \sum_{i=1}^{t-1} X_i Z_{i} )$.
    Therefore,
    \begin{align*}
        \Bigl\| \tilde{\theta}_{t-1} \Bigr\|_{V_{t-1}}
        & = \Bigl\| V_{t-1} \tilde{\theta}_{t-1} \Bigr\|_{V_{t-1}^{-1}}
        \\
        & = \Bigl\| W + \sum_{i=1}^{t-1} X_i Z_{t, i} \Bigr\|_{V_{t-1}^{-1}}
        \\
        & \le \left\| W \right\|_{V_{t-1}^{-1}} + \Bigl\| \sum_{i=1}^{t-1} X_i Z_{i} \Bigr\|_{V_{t-1}^{-1}}
        \, ,
    \end{align*}
    where the triangle inequality is used for the last inequality.
    To bound the first term, we use the fact that ${\lambda_{\max} ( V_{t-1}^{-1} ) \le \frac{1}{\lambda}}$, where $\lambda_{\max}( \cdot )$ denotes the maximum eigenvalue.
    It implies that $\| W \|_{V_{t-1}^{-1}} \le \frac{1}{\sqrt{\lambda}} \| W \|_2$.
    Since $\PP ( \| W \|_2 > L_{t, 0}^\delta ) \le \delta$ by assumption, $\| W \|_{V_{t-1}^{-1}} \le L_{t, 0}^\delta / \sqrt{\lambda}$ holds with probability at least $1 - \delta$. 
    The second term is bounded by \cref{lma:self-normalized bound}.
    With probability $1 - \delta$, it holds that
    \begin{equation*}
        \left\| \sum_{i=1}^{t-1} X_i Z_{t, i} \right\|_{V_{t-1}^{-1}} \le \sigma_R \sqrt{d \log \left( 1 + \frac{t}{d \lambda}\right) + 2 \log \frac{1}{\delta}}
        \, .
    \end{equation*}
    By taking the union bound over the two events, we obtain that
    \begin{equation*}
        \left\| \tilde{\theta}_{t-1} \right\|_{V_{t-1}} \le \frac{L_{t, 0}^\delta}{\sqrt{\lambda}} + \sigma_R \sqrt{d \log \left( 1 + \frac{t}{d \lambda}\right) + 2 \log \frac{1}{\delta} }
        \,
    \end{equation*}
    holds with probability at least $1 - 2\delta$, which proves the lemma.
\end{proof}

We also provide that the $\ell_2$-norm of the vector $W$ whose components are i.i.d. samples of a subGaussian distribution is upper-bounded with high probability.
This lemma, combined with \cref{lma:concentration lemma}, justifies $\Pcal_I$ to be a distribution over $\RR^d$ such that each component is an i.i.d. sample of a subGaussian distribution.

\begin{lemma}
\label{lma:bound 2norm}
    Suppose that $W \in \RR^d$ and each component of $W$ is sampled i.i.d. from a $\sigma_I^2$-subGaussian distribution.
    Take any $\delta \in (0, 1]$.
    Then, with probability $1 - \delta$, it holds that
    \begin{align*}
        \| W \|_2 \le \sigma_I \sqrt{ 2 d + 4 \log \frac{1}{\delta}}
        \, .
    \end{align*}
\end{lemma}
\begin{proof}
    Take $X_1 = \eb_1, \ldots, X_d = \eb_d$, where $\{ \eb_1, \ldots, \eb_d\} $ is the standard basis of $\RR^d$.
    By \cref{lma:self-normalized bound} with $\xi_i = (W)_i$ and $\lambda = 1$, it holds that
    \begin{align*}
        \left\| \sum_{i=1}^d (W)_i \eb_i \right\|_{ (2 I_d) ^{-1}} \le \sigma_I \sqrt{ d \log 2 + 2 \log \frac{1}{\delta}}
        \, 
    \end{align*}
    with probability at least $1 - \delta$.
    The proof is completed by noting that $\| \sum_{i=1}^d (W)_i \eb_i \|_{ (2 I_d) ^{-1}} = \frac{1}{\sqrt{2}} \| W \|_2$.
\end{proof}

Finally, we demonstrate that subGaussian distribution with lower-bounded variance satisfies the anti-concentration condition analogous to \cref{fact:gaussian}.
We normalize the distribution so that it is $1$-subGaussian.

\begin{lemma}[Sufficient condition for anti-concentration]
    Suppose that $\Pcal$ is a real-valued distribution that is symmetric, 1-subGaussian, and has variance at least $1/2$.
    Suppose $Z \in \RR^n$ for some $n \in \NN$ and its components are i.i.d. samples of $\Pcal$.
    Then, for any fixed $u \in \RR^n$, it holds that
    \begin{equation*}
        \PP \left( u^\top Z \ge \frac{1}{3} \| u \|_2 \right) \ge 0.01.
    \end{equation*}
\end{lemma}
\begin{proof}
    Without loss of generality, we may assume that $\left\| u \right\|_2 = 1$.
    Let $Y = u^\top Z$.
    Then, 
    $\Var(Y) = \Var(\sum_{i=1}^n (u)_i (Z)_i ) = \sum_{i=1}^n (u)_i^2 \Var((Z)_i) = \Var(\Pcal)\ge \frac{1}{2}$.
    On the other hand, we attain an upper bound of $\Var(Y)$ as follows:
    \begin{align}
        \Var(Y)
        & = \EE \left[ Y^2 \right] \nonumber
        \\
        & = \EE \left[ Y^2 \ind \left\{ |Y| \le \frac{1}{3} \right\} \right]
        + \EE \left[ Y^2 \ind \left\{ \frac{1}{3} < |Y| \le 4 \right\} \right]
        + \EE \left[ Y^2 \ind \left\{ |Y| \ge 4 \right\} \right] \nonumber
        \\
        & \le \frac{1}{9} \PP \left( |Y| \le \frac{1}{3} \right) + 16 \PP \left( \frac{1}{3} < |Y| \le 4 \right) 
        + \EE \left[ Y^2 \ind \left\{ |Y| \ge 4 \right\} \right] \nonumber
        \\
        & \le \frac{1}{9} + 16 \PP \left( \frac{1}{3} < |Y| \right) 
        + \EE \left[ Y^2 \ind \left\{ |Y| \ge 4 \right\} \right]
        \label{eq:upper bound of var(Y)}
        \, .        
    \end{align}
    We upper-bound $\EE [ Y^2 \ind \{ |Y| \ge 4 \} ]$ using its subGaussian property.
    Note that $Y = u^\top Z$ is 1-subGaussian since $\| u \|_2 = 1$ and the components of $Z$ are independent and 1-subGaussian.
    Applying the standard tail bound of subGaussian random variables, it holds that $\PP \left( |Y| \ge x \right) \le 2 \exp \left( - x^2 / 2 \right)$, or equivalently, $\PP \left( Y^2 \ge x \right) \le 2 \exp \left( - x / 2 \right)$.
    Then, it holds that
    \begin{align}
        \EE \left[ Y^2 \ind \left\{ Y^2 \ge 16 \right\} \right]
        & = \int_{0}^\infty \PP \left( Y^2 \ind \left\{ Y^2 \ge 16 \right\} \ge x \right) \, dx \nonumber
        \\
        & = \int_{0}^{16}\PP \left( Y^2 \ind \left\{ Y^2 \ge 16 \right\} \ge x \right) \, dx + \int_{16}^\infty \PP \left( Y^2 \ind \left\{ Y^2 \ge 16 \right\} \ge x \right) \, dx \nonumber
        \\
        & = 16 \PP \left( Y^2 \ge 16 \right) + \int_{16}^\infty \PP \left( Y^2 \ge x \right) \, dx \nonumber
        \\
        & \le 32 e^{-8} + \int_{16}^\infty 2 e^{-\frac{x}{2}} \, dx \nonumber
        \\
        & = 32 e^{-8} + 4 e^{-8} \nonumber
        \\
        & \le 0.0121
        \label{eq:upper bound of Y^2}
        \, .
    \end{align}
    By plugging in the upper bound of~\eqref{eq:upper bound of Y^2} to inequality~\eqref{eq:upper bound of var(Y)} and reordering the terms, we obtain that
    \begin{align*}
        \PP \left( |Y| > \frac{1}{3} \right) \ge \frac{1}{16} \left( \frac{1}{2} - \frac{1}{9} - 0.0121 \right) \ge 0.023
        \, .
    \end{align*}
    Finally, recall that $Z$ is symmetric, therefore $Y$ is symmetric.
    Therefore, we have $\PP \left( Y \ge \frac{1}{3} \right) \ge \frac{0.023}{2} \ge 0.01$.
\end{proof}

\section{Auxiliary Lemmas}

\begin{lemma}[Azuma-Hoeffding Inequality]
\label{lma:Azuma-Hoeffding inequality}
    Fix $n \in \NN$.
    Let $\left\{ Z_i \right\}_{i=1}^n$ be a sequence of real-valued random variables adapted to a filtration $\left\{ \Fcal_i \right\}_{i=0}^n$.
    Suppose that there exists $a < b$ such that $Z_i \in [a, b]$ holds almost surely for all $i \in [n]$.
    Then, for any $\delta \in (0, 1]$, the following inequality holds with probability at least $1 - \delta$:
    \begin{equation*}
        \sum_{i=1}^n \left( Z_i - \EE \left[ Z_i \mid \Fcal_{t-1} \right] \right) \le (b - a) \sqrt{ \frac{n}{2} \log \frac{1}{\delta}}
        \, .
    \end{equation*}
\end{lemma}

\begin{lemma}[Lemma~1 of~\citet{laurent2000adaptive}]
\label{lma:chi-squared}
    Let $Y_1, \ldots, Y_d$ be i.i.d. standard Gaussian variables.
    Set $Z = \sum_{i=1}^d (Y_i^2 - 1)$.
    Then, the following inequality holds for any $x > 0$,
    \begin{equation*}
        \PP( Z \ge \sqrt{ d x} + 2 x ) \le e^{-x}
        \, .
    \end{equation*}
\end{lemma}


\end{document}